\documentclass[10pt,twocolumn,letterpaper]{article}

\usepackage{iccv}
\usepackage{times}
\usepackage{epsfig}
\usepackage{graphicx}
\usepackage{amsmath}
\usepackage{amssymb}

\usepackage{bbm}
\usepackage{booktabs,dcolumn}
\usepackage{multirow}

\usepackage{color}
\newcommand{\ywunew}[1]{\textcolor{black}{#1}}
\newcommand{\ywu}[1]{\textcolor{black}{#1}}

\usepackage[pagebackref=true,breaklinks=true,letterpaper=true,colorlinks,bookmarks=false]{hyperref}

\iccvfinalcopy 

\def\iccvPaperID{6144} 

\ificcvfinal\fi

\begin{document}

\title{Rethinking Counting and Localization in Crowds: \\A Purely Point-Based Framework}
\author{Qingyu Song\textsuperscript{\rm 1}\thanks{\textit{Equal contribution.} $^\dagger$\textit{Corresponding author.}}\hspace{1em}Changan Wang\textsuperscript{\rm 1}$^*$\hspace{1em}Zhengkai Jiang\textsuperscript{\rm 1}\hspace{1em}Yabiao Wang\textsuperscript{\rm 1}\\Ying Tai\textsuperscript{\rm 1}\hspace{1em}Chengjie Wang\textsuperscript{\rm 1}\hspace{1em}Jilin Li\textsuperscript{\rm 1}\hspace{1em}Feiyue Huang\textsuperscript{\rm 1}$^\dagger$\hspace{1em}Yang Wu\textsuperscript{\rm 2}\hspace{1em}\\
\textsuperscript{\rm 1}Tencent Youtu Lab, \textsuperscript{\rm 2}Applied Research Center (ARC), Tencent PCG\\
{\tt\small qingyusong@zju.edu.cn, \{changanwang, zhengkjiang, caseywang\}@tencent.com}\\{\tt\small \{yingtai, jasoncjwang, jerolinli, garyhuang, dylanywu\}@tencent.com}
}

\maketitle
\ificcvfinal\thispagestyle{empty}\fi

\begin{abstract}
     Localizing individuals in crowds \ywunew{is more in accordance with the practical demands of subsequent high-level crowd analysis tasks than simply counting}. However, existing localization based methods \ywunew{relying} on intermediate representations (\textit{i.e.}, density maps or pseudo boxes) \ywunew{serving as} learning targets \ywunew{are counter-intuitive} and error-prone. In this paper, we propose a purely point-based framework for joint crowd counting and individual localization. \ywunew{For this framework}, instead of merely reporting the absolute counting error at image level, we propose a new metric, called density Normalized Average Precision (nAP), to provide \ywunew{more comprehensive and more precise performance evaluation}. Moreover, \ywunew{we design an intuitive solution under this framework, which is called} Point to Point Network (P2PNet). \ywunew{P2PNet} discards superfluous steps and directly predicts a set of point proposals to represent heads in an image, being consistent with the human annotation \ywunew{results}. By thorough analysis, we reveal the key step towards implementing such a novel idea is to assign optimal learning targets for these proposals. Therefore, we propose to conduct this crucial association in an one-to-one matching manner using the Hungarian algorithm. The P2PNet not only significantly surpasses state-of-the-art methods on popular counting benchmarks, but also achieves promising localization accuracy. The codes will be available at: \href{https://github.com/TencentYoutuResearch/CrowdCounting-P2PNet}{TencentYoutuResearch/CrowdCounting-P2PNet}.
\end{abstract}

\section{Introduction}
\begin{figure}[t!]
    \centering
    \includegraphics[width=1\columnwidth]{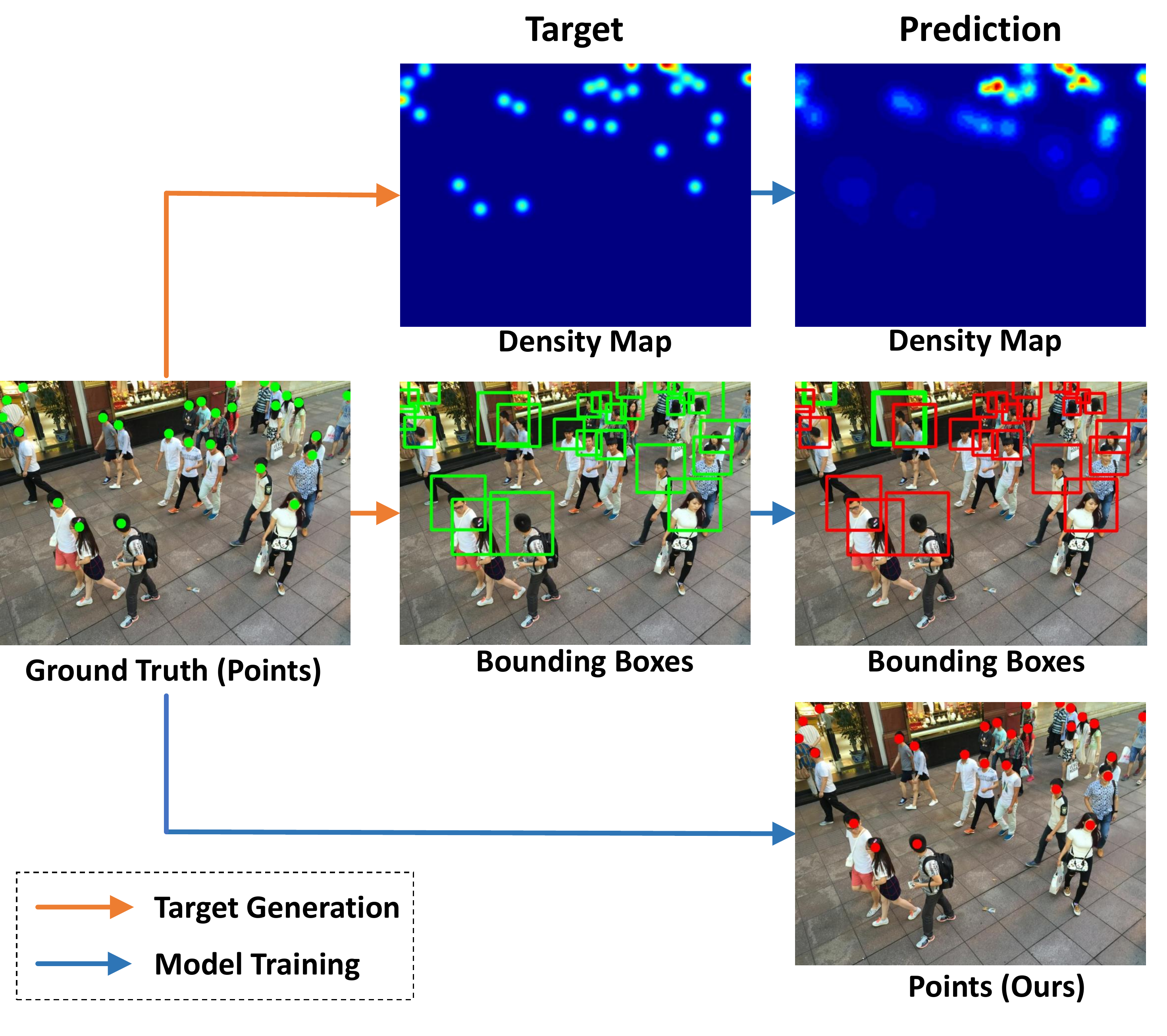}
    \caption{Illustrations for the comparison of our pipeline with existing methods, in which the predictions are marked in Red while the ground truths are marked as Green. \textbf{Top flow:} The dominated density map learning based methods fail to provide the exact locations of individuals. \textbf{Middle flow:} The \ywu{estimated inaccurate} ground truth bounding boxes \ywu{make} the detection based methods error-prone, such as the missing detections as indicated, especially for the NMS-like process. \textbf{Bottom flow:} Our pipeline directly predicts a set of points to represent the locations of individuals, which is simple, intuitive and competitive as demonstrated, bypassing those error-prone steps. Best viewed in color.}
    \label{fig1}
    \vspace{-1.8em}
\end{figure}
Among all the \ywu{related concrete} tasks of crowd analysis, crowd counting \ywu{is a} fundamental pillar, aiming to estimate the number of individuals in a crowd. However, simply giving a \ywu{single number} is obviously far from \ywu{being able to support} the practical demands of subsequent \ywu{higher}-level crowd analysis \ywu{tasks}, such as crowd tracking, \ywu{activity recognition}, abnormality detection, \ywu{flow/behavior prediction, \textit{etc}}. 

\ywu{In fact}, there is an obvious tendency in this field for more challenging fine-grained estimation (\textit{i.e.}, the locations of individuals) beyond simply counting. Specifically, some approaches cast crowd counting as a head detection problem, but leaving more efforts on labor-intensive \ywunew{annotation} for tiny-scale heads. Other approaches \cite{liu2019point, sam2020locate} attempted to generate the pseudo bounding boxes of heads with only point annotations provided, which however appears to be tricky or inaccurate at least. Also trying to directly locate individuals, several methods \cite{laradji2018blobs,liu2019recurrent} got stuck in suppressing or splitting over-close \ywunew{instance candidates}, making themselves error-prone due to the extreme head scale variation, especially for highly-congested regions. To eschew the above problems, we propose a purely point-based framework for \ywunew{jointly} counting and localizing individuals in crowds. This framework directly uses point annotations as learning targets and simultaneously outputs points to locate individuals, benefiting from the high-precision localization property of point representation and its relatively cheaper annotation cost. The pipeline is illustrated in Figure \ref{fig1}.

Additionally, in terms of the evaluation metrics, some farsighted works \cite{guerrero2015extremely, tian2019padnet} encourage to adopt patch-level metrics for fine-grained evaluation, but \ywunew{they only provide a rough measure for localization}. Other existing localization aware metrics either ignore the significant density variation across crowds \cite{liu2019point,sam2020locate} or lack the punishment for duplicate predictions \cite{sam2020locate,wang2020nwpu}. Instead, we propose a new metric called density Normalized Average Precision (nAP) to provide a comprehensive evaluation metric for both localization and counting errors. The nAP metric supports both box and point representation as inputs (\textit{i.e.}, predictions or annotations), without the defects mentioned above.


Finally, as an intuitive solution under this new framework, we develop a novel method to directly predict a set of point proposals with the coordinates of heads in an image and their confidences. Specifically, we propose a \ywunew{Point-to-Point Network (P2PNet)} to directly receive a set of annotated head points for training and predict \ywunew{points too during inference}. Then to make such an idea \ywunew{work} correctly, we delve into the ground truth \ywunew{target} assignation process to reveal the crucial of such association. The conclusion is that either the case when \textit{multiple} proposals are matched to \ywunew{a} \textit{single} ground truth, or the opposite case, \ywunew{can make} the model confused during training, leading to over-estimated or under-estimated \ywu{counts}. So we propose to perform an one-to-one matching by Hungarian algorithm to associate the point proposals with their ground truth targets, and \ywunew{the} unmatched proposals should be classified as negatives. We empirically show that such \ywunew{a} matching is beneficial to \ywunew{improving} the nAP metric, serving as a key component for our solution under the new framework. This simple, intuitive and efficient design yields state-of-the-art counting performance and promising localization accuracy. 

The \ywunew{major} contributions of this \ywunew{work} are \ywunew{three-fold}:

1. We propose a purely point-based framework for joint counting and \ywunew{individual localization} in crowds. This framework encourages fine-grained predictions, benefiting the practical demands of downstream tasks in crowd analysis.

2. We propose a new metric termed density Normalized Average Precision to account for \ywunew{the evaluation of} both localization \ywunew{and counting}, as a comprehensive evaluation metric under the new framework.

3. We propose P2PNet as an intuitive solution following this conceptually simple framework. The method achieves state-of-the-art counting accuracy and promising localization performance, and might also be inspiring for other tasks relying on point predictions.

\section{Related Works}
In this section, we review two kinds of crowd counting methods in recent literature\ywu{. They} are grouped according to whether locations of individuals could be provided. Since we focus on the estimation of locations, existing metrics accounting for localization errors are also discussed.
\vspace{-1.0em}
\paragraph{Density Map based Methods.}The adoption of density map \ywu{is a common choice of} most state-of-the-art \ywu{crowd} counting methods, since it was firstly introduced in \cite{lempitsky2010learning}. And the estimated count is obtained by summing over the predicted density maps. Recently, many efforts have been devoted to \ywu{pushing} forward the counting performance frontier of such methods. They either conduct a pixel-wise density map regression \cite{li2018csrnet,miao2020shallow,jiang2020attention,bai2020adaptive,liu2020adaptive,hu2020count}, or resort to classify the count value of local patch into several bins \cite{xiong2019open,liu2019counting,liu2020weighing}. Although many compelling models have been proposed, these density map learning based models still fail to provide the exact locations of individuals in crowds, not to mention their inherent flaws as pointed out in \cite{bai2020adaptive,ma2019bayesian,liu2019counting}. Whereas the proposed method goes beyond counting and focuses on the direct prediction for locations of individuals, eschewing the defects of density maps and also benefiting the downstream practical applications.
\vspace{-1.0em}
\paragraph{Localization based Methods.}These methods typically achieve counting by firstly predicting the locations of individuals. Motivating by cutting-edge object detectors, some counting methods \cite{lian2019density,liu2019point,sam2020locate} try to predict the bounding boxes for heads of individuals. However, with only the point annotations available, these methods rely on heuristic estimation for ground truth bounding boxes, which is error-prone or even infeasible. These inaccurate bounding boxes not only confuse the model training process, but also make the post-process, \textit{i.e.}, NMS, fail to suppress false detections. Without those inaccurate targets introduced, other methods locate individuals by points \cite{liu2019recurrent} or blobs \cite{laradji2018blobs}, but leaving more efforts to remove duplicates or split over-close detected individuals in congest regions. Instead, bypassing these tricky post-processing with an one-to-one matching, we propose to streamline the framework to directly estimate the point locations of individuals.
\vspace{-1.0em}
\paragraph{Localization Aware Metrics.}Traditional universally agreed evaluation metrics only measure the counting errors, entirely ignoring the significant spatial variation of estimation errors in single image. To provide a more accurate evaluation, some works \cite{guerrero2015extremely,liu2019geometric, tian2019padnet} advocate to adopt patch-level or pixel-level absolute counting error as criteria, in lieu of the commonly used image-level metric. Other research \cite{sam2020locate} proposes Mean Localization Error to compute the average pixel distance between the predictions and ground truths, merely evaluating the localization errors. Inspired by evaluation metric used in object detection, \cite{idrees2018composition} proposes to use the area under the Precision-Recall curve after a greedy association, which however ignores the punishment for duplicate predictions. Hence, \cite{liu2019recurrent} proposes to adopt a sequential matching and then use the standard Average Precision (AP) for evaluation. In this paper, we propose a new metric, termed density Normalized Average Precision (nAP), as a comprehensive evaluation metric for both localization errors and false detections. In particular, the nAP metric introduces a density normalization to account for the large density variation problem in crowds.
\section{Our Work}
We firstly introduce the proposed framework in detail (Sec. \ref{label_task}), and the new evaluation metric nAP is also presented (Sec. \ref{label_metric}). Then we conduct a thorough analysis to reveal the key issue in improving the nAP metric under the new framework (Sec. \ref{label_work}). Inspired by the insightful analysis, we introduce the proposed P2PNet (Sec. \ref{label_p2p}), which directly predicts a set of point proposals to represent heads.
\subsection{The Purely Point-based Framework}\label{label_task}
The proposed framework directly receives point annotations as its learning targets and then provides the exact locations for individuals in a crowd, rather than simply counting the number of individuals within it. And the locations of individuals are typically indicated by the center points of heads, possibly with optional confidence scores. 

Formally, given an image with $N$ individuals, we use $p_i=(x_i,y_i)$, $i \in \{1,..,N\}$, to represent the head's center point of the $i$-th individual, which is located in $(x_i,y_i)$. Then the collection of the center points for all individuals could be further denoted as $\mathcal{P}=\left\{p_i | i \in \{1,..,N\}\right\}$. Assuming a well-designed model $\mathcal{M}$ is trained to instantiate this new framework. And the model $\mathcal{M}$ predicts another two collections $\hat{\mathcal{P}}=\left\{\hat{p}_j|j \in \{1,..,M\}\right\}$ and $\hat{\mathcal{C}}=\left\{\hat{c}_j | j \in \{1,..,M\}\right\}$, in which $M$ is the number of predicted individuals, and $\hat{c_j}$ is the confidence score of the predicted point $\hat{p}_j$. Without loss of generality, we may assume that $\hat{p}_j$ is exactly the prediction for the ground truth point $p_i$. \textit{Then our goal is to ensure that the distance between $\hat{p}_j$ and $p_i$ is as close as possible with a sufficiently high score $\hat{c_j}$.} As a byproduct, the number of predicted individuals $M$ should also be close enough to the ground truth crowd number $N$. In a nutshell, the new framework could simultaneously achieve crowd counting and individual localization.

Compared with traditional counting methods, the individual locations provided by this framework are helpful to those motion based crowd analysis tasks, such as crowd tracking \cite{zhu2016crowd}, activity recognition \cite{dupont2017crowd}, abnormality detection \cite{chen2019detecting}, etc. Besides, without relying on labor-intensive annotations, inaccurate pseudo boxes or tricky post-processing, this framework benefits from the high-precision localization property of original point representation, especially for highly-congested regions in crowds.

Therefore, this new framework is worth more attentions due to its advantages and practical values over traditional crowd counting. However, since the existence of severe occlusions, density variations, and annotation errors, it is quite challenging to tackle with such a task \cite{liu2019recurrent,liu2019point,sam2020locate}, which even is considered as ideal but infeasible in \cite{idrees2018composition}. 
\subsection{Density Normalized Average Precision}\label{label_metric}
It is natural to ask that how to evaluate the performance of model $\mathcal{M}$ under the above new framework. In fact, a well-performed model following this framework should not only produce as few as false positives or false negatives, but also achieve competitive localization accuracy. Therefore, motivated by the mean Average Precision (mAP) \cite{lin2014microsoft} metric widely used in Object Detection, we propose a density Normalized Average Precision (nAP) to evaluate both the localization errors and counting performance. 

The nAP is calculated based on the Average Precision, which is the area under the Precision-Recall (PR) curve. And the PR curve could be easily obtained by accumulating a binary list following the common practice in \cite{lin2014microsoft}. In the binary list, a True Positive (TP) prediction is indicated by 1, and a False Positive (FP) prediction is indicated by 0. Specifically, given all predicted head points $\hat{\mathcal{P}}$, we firstly sort the point list with their confidence scores from high to low. Then we sequentially determine that the point under investigation is either TP or FP, according to a pre-defined density aware criterion. Different from the greedy association used in \cite{idrees2018composition,sam2020locate}, we apply a sequential association in which those higher scored predictions are associated firstly. In this way, these TP predictions could be easily obtained by a simple threshold filtering during inference.

\begin{figure}[t!]
    \centering
    \includegraphics[width=0.9\columnwidth]{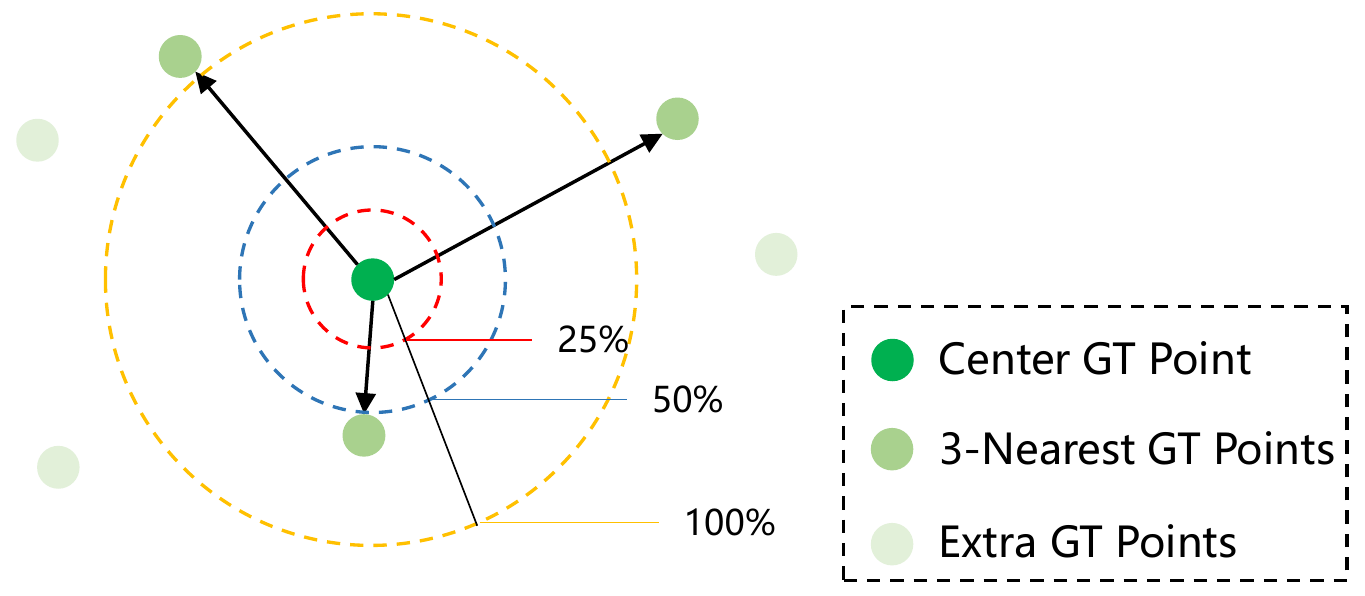}
    \caption{Illustration for different levels of localization accuracy in nAP ($k$=3). The yellow circle indicates the region within $d_{k\text{NN}}(p_i)$ pixels from the center GT point $p_i$. A typical value for $\delta$ is 0.5, as indicated by the blue circle, which means that the nearest GT point of most pixels within this region should be $p_i$. The red circle represents a threshold ($\delta$=0.25) for \ywu{stricter} localization accuracy.}
    \label{fig2}
    \vspace{-1.2em}
\end{figure}
We introduce our density aware criterion as follows. A predicted point $\hat{p}_j$ is classified as TP only if it could be matched to certain ground truth $p_i$, in which $p_i$ must not be matched before by any higher-ranked point. The matching process is guided by a pixel-level Euclidean distance based criterion $\mathbbm{1}(\hat{p}_j, p_i)$. However, directly using the pixel distance to measure the affinity ignores the side effects from the large density variation across crowds. Thus, we introduce a density normalization for this matching criterion to mitigate the density variation problem. The density around a certain ground truth point is estimated following \cite{zhang2016single}. Formally, the final criterion used in nAP is defined as:
\begin{equation}
    \mathbbm{1}(\hat{p}_j, p_i) =
    \left\{\begin{aligned}
            1 & , & \text{if}\ d(\hat{p}_j, p_i) / d_{k\text{NN}}(p_i) < \delta, \\
            0 & , & \text{otherwise}, \qquad\qquad\quad 
    \end{aligned} \right.
\end{equation} %
where $d(\hat{p}_j, p_i)=||\hat{p}_j - p_i||_{2}$ denotes to the Euclidean distance, and $d_{k\text{NN}}(p_i)$ denotes the average distance to the $k$ nearest neighbors of $p_i$. We use a threshold $\delta$ to control the desired localization accuracy, as shown in Figure \ref{fig2}.
\subsection{Our Approach}\label{label_work}
Our approach is an intuitive solution following the proposed framework, which directly predict a set of point proposals to represent the center points for heads of individuals. In fact, the idea of point prediction is not new to the vision community, although it is quite different here. To name a few, in the field of pose estimation, some methods adopt heatmap regression \cite{chen2018cascaded,xiao2018simple} or direct point regression \cite{toshev2014deeppose,xiong2019a2j} to predict the locations of pre-defined keypoints. Since the number of the keypoints to be predicted is fixed, the learning targets for these point proposals could be determined entirely before the training. Differently, the proposed framework aims to predict a point set of unknown size and is an open-set problem by nature \cite{xiong2019open}. Thus, one crucial problem of such a methodology is to determine which ground truth point should the current prediction be responsible for.

\begin{figure}[t!]
    \centering
    \includegraphics[width=1\columnwidth]{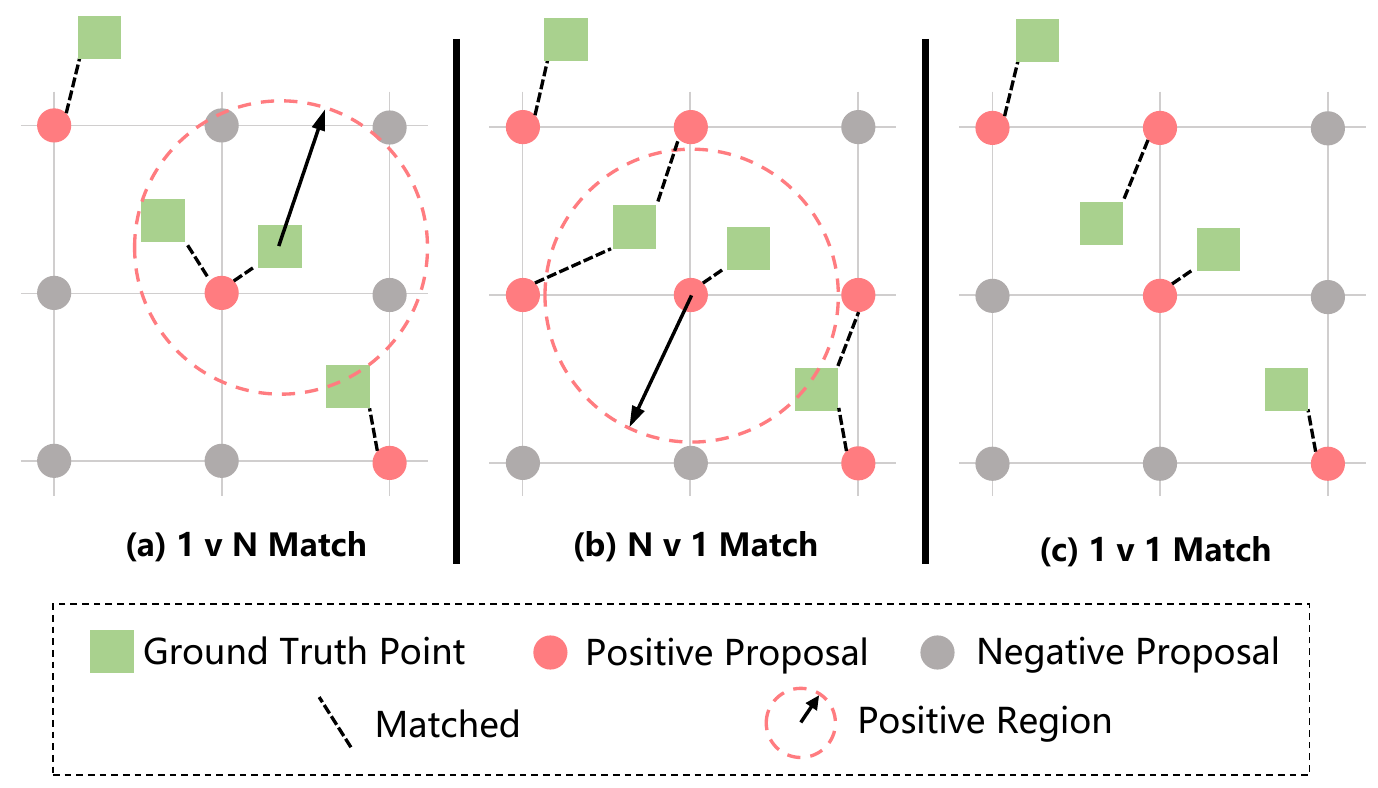}
    \caption{(a) Multiple ground truth points might be matched to the same proposal when selecting the nearest proposal for each of them, which leads to under-estimated \ywu{counts}. (b) Multiple proposals might be matched to the same ground truth point when selecting the nearest ground truth for each of them, which leads to over-estimated \ywu{counts}. (c) Our One-to-One match is without the above two defects, thus is suitable for direct point prediction. }
    \label{fig3}
    \vspace{-1.5em}
\end{figure}

We propose to solve this key problem with a mutually optimal one-to-one association strategy during the training stage. Let us conduct a thorough analysis to show the defects of the other two strategies for the ground truth targets assignment. Firstly, for each ground truth point, the proposal with the nearest distance should produce the best prediction. However, if we select the nearest proposal for every ground truth point, it is likely that one proposal might be matched to multiple ground truth points, as shown in Figure \ref{fig3} (a). In such a case, only one ground truth could be correctly predicted, leading to under-estimated \ywu{counts}, especially for the congested regions. Secondly, for each point proposal, we may assign the nearest ground truth point as its target. Intuitively, this strategy might be helpful to alleviate the overall overhead of the optimization, since the nearest ground truth point is relatively easier to predict. However, in such an assignment, there may exist multiple proposals which simultaneously predict the same ground truth, as shown in Figure \ref{fig3} (b). Because there are no scale annotations available, it is tricky to suppress these duplicate predictions, which might lead to over-estimated. 
Consequently, the association process should take both sides into consideration and produces the mutually optimal one-to-one matching results, as shown in Figure \ref{fig3} (c).

Additionally, both the other two strategies have to determine a negative threshold, and the proposals whose distance with their matched targets are above this threshold will be considered as negatives. While using the one-to-one matching, those unmatched proposals are automatically remained as negatives, without any hyperparameter introduced. \textit{In a nutshell, the key to solve the open-set direct point prediction problem is to ensure a mutually optimal one-to-one matching between predicted and ground truth points.} 

After the ground truth targets are obtained, these point proposals could be trained through an end-to-end optimization. Finally, the positive proposals should be pushed toward their targets, while those negative proposals would be simply classified as backgrounds. Since the point proposals are dynamically updated along with the training process, those proposals which have the potential to perform better could be gradually selected by the one-to-one matching to serve as the final predictions.

Actually, the distance used in above matching could be any other cost measure beyond pixel distance, such as a combination of confidence score and pixel distance. We empirically show that taking confidence scores of proposals into consideration during the one-to-one matching is helpful to improve the proposed nAP metric. Let us consider two predicted proposals around the same ground truth point $p_i$. If they have the same confidence score, the one closer to $p_i$ should be matched as positive and encouraged to achieve higher localization accuracy. While the other one proposal should be matched as negative and supervised to lower its confidence, thus might not be matched again during next training iteration. On the contrary, if the two proposals share the same distance from $p_i$, the one with higher confidence should be trained to be closer to $p_i$ with a much \ywu{higher} confidence. Both the above two cases would encourage the positive proposals to have more accurate locations as well as relatively higher confidences, which is beneficial to the improvement of nAP under the proposed framework. 
\subsection{The P2PNet Model}\label{label_p2p}
\begin{figure}[t!]
    \centering
    \includegraphics[width=1\columnwidth]{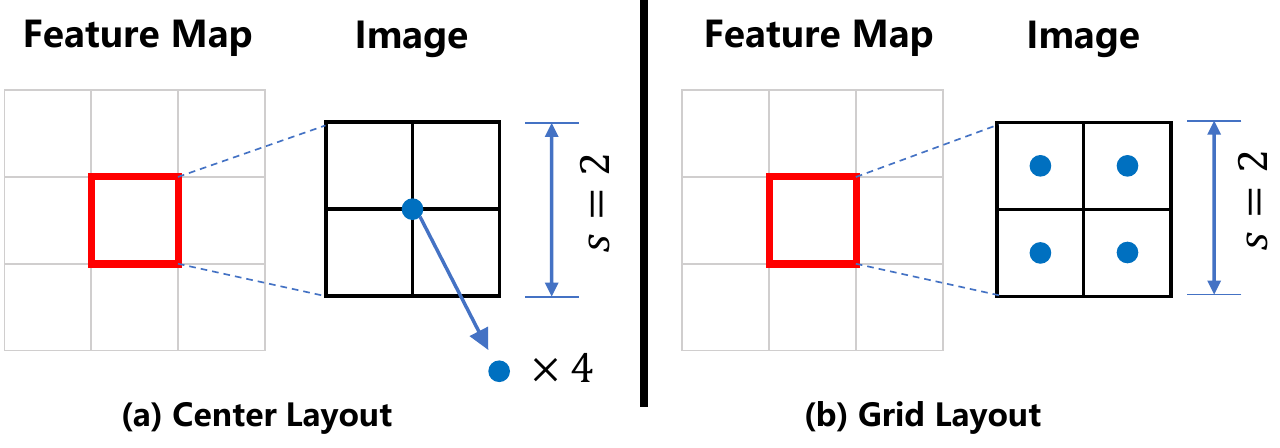}
    \caption{Two types of layout for reference points ($s = 2$, $K=4$).}
    \label{fig4}
    \vspace{-1.0em}
\end{figure}
\begin{figure*}[t!] 
  \centering
  \includegraphics[width=0.95\textwidth]{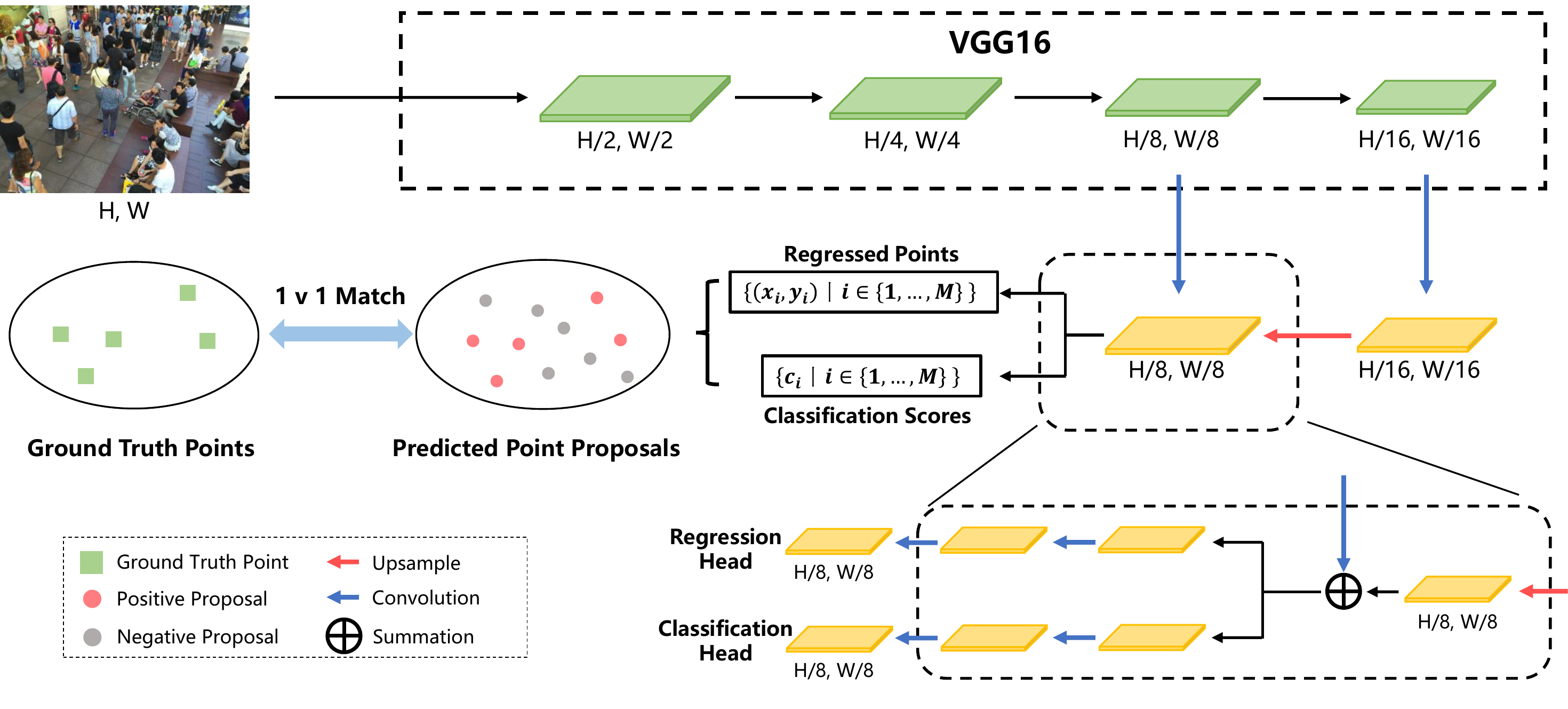}
  \vspace{-0.8em}
  \caption{The overall architecture of the proposed P2PNet. Built upon the VGG16, it firstly introduce an upsampling path to obtain fine-grained deep feature map. Then it exploits two branches to simultaneously predict a set of point proposals and their confidence scores. The key step in our pipeline is to ensure an one-to-one matching between point proposals and ground truth points, which determines the learning targets of those proposals.}
  \label{fig5} 
  \vspace{-1.5em}
\end{figure*}
In this part, we present the detailed pipeline of the proposed Point to Point Network (P2PNet). Begining with the generation of point proposals, we introduce our one-to-one association strategy in detail. Then we present the loss function and the network architecture for the P2PNet.
\vspace{-1.0em}
\paragraph{Point Proposal Prediction.} Let us denote the deep feature map outputted from the backbone network by $\mathcal{F}_s$, in which $s$ is the downsampling stride and $\mathcal{F}_s$ is with a size of $H\times W$. Then based on $\mathcal{F}_s$, we adopt two parallel branches for point coordinate regression and proposal classification. For the classification branch, it outputs the confidence scores with a Softmax normalization. For the regression branch, it resorts to predict the offsets of the point coordinates due to the intrinsic translation invariant property of convolution layers. Specifically, each pixel on $\mathcal{F}_s$ should correspond to a patch of size $s \times s$ in the input image. In that patch, we firstly introduces a set of fixed reference points $\mathcal{R}=\{R_k | k \in \{1,...,K\}\}$ with pre-defined locations $R_k=(x_k,y_k)$. These reference points could be either densely arranged on the patch or just set to the center of that patch, as shown in Figure \ref{fig4}. Since there are $K$ reference points for each location on $\mathcal{F}_s$, the regression branch should produce totally $H\times W \times K$ point proposals. Assuming the reference point $R_k$ predicts offsets $(\Delta_{jx}^k,\Delta_{jy}^k)$ for its point proposal $\hat{p}_j=(\hat{x}_j,\hat{y}_j)$, then the coordinate of $\hat{p}_j$ is calculated as follows:
\begin{equation}
    \begin{aligned}
            \hat{x}_j = x_k + \gamma \Delta_{jx}^k, \\
            \hat{y}_j = y_k + \gamma \Delta_{jy}^k,
    \end{aligned} 
\end{equation} 
where $\gamma$ is a normalization term, which scales the offsets to rectify the relatively small predictions.
\vspace{-0.8em}
\paragraph{Proposal Matching.} Following the symbols defined in Sec. \ref{label_task}, we assign the ground truth target from $\hat{\mathcal{P}}$ for every point proposal in $\mathcal{P}$ using an one-to-one matching strategy $\Omega(\mathcal{P}, \hat{\mathcal{P}}, \mathcal{D})$. The $\mathcal{D}$ is a pair-wise matching cost matrix with the shape $N\times M$, which measures the distance between two points in a pair. Instead of simply using the pixel distance, we also consider the confidence score of that proposal, since we encourage the positive proposals to have higher confidences. Formally, the cost matrix $\mathcal{D}$ is defined as follows:
\begin{equation}
    \mathcal{D}(\mathcal{P}, \hat{\mathcal{P}}) = \left( \tau \left|\left|p_i - \hat{p}_j\right|\right|_{2} - \hat{c}_j\right)_{i\in N, j\in M},
\end{equation} 
where $\left|\left|\cdot \right|\right|_{2}$ denotes to the $l_2$ distance, and $\hat{c}_j$ is the confidence score of the proposal $\hat{p}_j$. $\tau$ is a weight term to balance the effect from the pixel distance.

Based on the pair-wise cost matrix $\mathcal{D}$, we conduct the association using the Hungarian algorithm \cite{kuhn1955hungarian,e2e2016,s2s2019} as the matching strategy $\Omega$. Note that in our implementation, we ensure $M>N$ to produce many enough predictions, since those redundant proposals would be classified as negatives. From the perspective of the ground truth points, let us use a permutation $\xi$ of $\{1,...,M\}$ to represent the optimal matching result, \textit{i.e.}, $\xi = \Omega (\mathcal{P}, \hat{\mathcal{P}}, \mathcal{D})$. That is to say, the ground truth point $p_i$ is matched to the proposal $\hat{p}_{\xi(i)}$. Furthermore, those matched proposals (positives) could be represented as a set $\hat{\mathcal{P}}_{pos} = \{\hat{p}_{\xi(i)} | i\in\left\{1,...,N\}\right\}$, and those unmatched proposals in the set $\hat{\mathcal{P}}_{neg} = \left\{\hat{p}_{\xi(i)} | i\in\{N+1,...,M\}\right\}$ are labeled as negatives.
\vspace{-0.8em}
\paragraph{Loss Design.} After the ground truth targets have been obtained, we calculate the Euclidean loss $\mathcal{L}_{loc}$ to supervise the point regression, and use Cross Entropy loss $\mathcal{L}_{cls}$ to train the proposal classification. The final loss function $\mathcal{L}$ is the summation of the above two losses, which is defined as:
\begin{equation}
\mathcal{L}_{cls} = -\frac{1}{M} \left\{ \sum_{i=1}^N \mathrm{log}\;\hat{c}_{\xi(i)}
 + \lambda_1 \sum_{i=N+1}^M \mathrm{log}\left(1 - \hat{c}_{\xi(i)}\right) \right\},
\end{equation}
\begin{equation}
\mathcal{L}_{loc} = \frac{1}{N} \sum_{i=1}^N \left|\left|p_i - \hat{p}_{\xi(i)}\right|\right|_{2}^2,
\end{equation}
\begin{equation}
    \mathcal{L} = \mathcal{L}_{cls} + \lambda_2 \mathcal{L}_{loc},
\end{equation} 
where $\left|\left|\cdot \right|\right|_{l_2}$ denotes to the Euclidean distance, $\lambda_1$ is a re-weight factor for negative proposals, and $\lambda_2$ is a weight term to balance the effect of the regression loss.
\vspace{-0.8em}
\paragraph{Network Design.} As illustrated in Figure \ref{fig5}, we use the first 13 convolutional layers in VGG-16\_bn \cite{simonyan2014very} to extract deep features. With the outputted feature map, we upsample its spatial resolution by a factor of 2 using nearest neighbor interpolation. Then the upsampled map is merged with the feature map from a lateral connection by element-wise addition. This lateral connection is used to reduce channel dimensions of the feature map after the fourth convolutional blcok. Finally, the merged feature map undergoes a $3\times 3$ convolutional layer to get $\mathcal{F}_s$, and the convolution in which is used to reduce the aliasing effect due to the upsampling.

The prediction head in our P2PNet is consisted of two branches, which are both fed with $\mathcal{F}_s$ and produce point locations and confidence scores respectively. For simplicity, the architecture of the two branches are kept same, which is consisted of three stacked convolutions interleaved with ReLU activations. We have empirically found this simple structure yield competitive results.
\section{Experiments}
\subsection{Implementation Details}
\paragraph{Dataset.} We exploit existing publicly available datasets in crowd counting to demonstrate the superiority of our method. Specifically, extensive experiments are conducted on four challenging datasets, including ShanghaiTech PartA and PartB \cite{zhang2016single}, UCF\_CC\_50 \cite{idrees2013multi}, UCF-QNRF \cite{idrees2018composition} and NWPU-Crowd \cite{wang2020nwpu}. For experiments on UCF\_CC\_50, we conduct a five-fold cross validation following \cite{idrees2013multi}.

\vspace{-1.0em}
\paragraph{Data Augmentations.} We firstly adopt random scaling with its scaling factor selected from [0.7, 1.3], keeping the shorter side not less than 128. Then we randomly crop an image patches with a fixed-size of $128\times 128$ from the resized image. Finally, random flipping with a probability of 0.5 is also adopted. For the datasets containing extremely large resolution, \textit{i.e.}, QNRF and NWPU-Crowd, we keep the max size of image no longer than 1408 and 1920, respectively, and keep the original aspect ratio.
\vspace{-1.0em}
\paragraph{Hyperparameters.} We use the feature map of stride $s=8$ for the prediction. The number $K$ of the reference points is set to 4 (8 for QNRF dataset). And $K$ is set according to the dataset statistics to ensure $M>N$. For the point regression, we set the $\gamma$ to 100. The weight term $\tau$ during the matching is set as 5e-2. In the loss function, the $\lambda_1$ is set to 0.5, and $\lambda_2$ is set to 2e-4. Adam algorithm \cite{kingma2014adam} with a fixed learning rate 1e-4 is used to optimize the model parameters. Since the weights in the backbone network have been pre-trained on the ImageNet, thus, we use a smaller learning rate 1e-5. The training batch size is set to 8.

\subsection{Model Evaluation}
\begin{table*}[ht!]
    \vspace{-0.5em}
    \centering
    \small{
    \resizebox{0.9\linewidth}{!}{
    \begin{tabular}{c|c|c|c|c|c}
        \toprule[1pt]
        nAP$_{\delta} $&SHTech PartA& SHTech PartB & UCF\_CC\_50 & UCF-QNRF & NWPU-Crowd \\
        \midrule[0.5pt]
        $\delta=0.05$ & 10.9\% & 23.8\%  & 5.0\% & 5.9\% & 12.9\% \\  
        $\delta=0.25$ & 70.3\% & 84.2\%  & 54.5\% & 55.4\% & 71.3\% \\  
        $\delta=0.50$ & 90.1\% & 94.1\%  & 88.1\% & 83.2\% & 89.1\% \\  
        $\delta=\{{0.05:0.05:0.50}\}$ & 64.4\% & 76.3\%  & 54.3\% & 53.1\% & 65.0\% \\  
        \bottomrule[1pt]
    \end{tabular}}
    }\vspace{0.2em}
    \caption{The overall performance of our P2PNet.}
    \label{tab:loc}
    \vspace{-0.4em}
\end{table*}
\begin{table*}[ht!]
    \vspace{-0.6em}
    \centering
    \small{
    \resizebox{0.9\linewidth}{!}{
    \begin{tabular}{c|c|cc|cc|cc|cc}
        \toprule[1pt]
        \multirow{2}{*}{Methods} & \multirow{2}{*}{Venue} & \multicolumn{2}{c|}{SHTech PartA} & \multicolumn{2}{c|}{SHTech PartB} & \multicolumn{2}{c|}{UCF\_CC\_50} & \multicolumn{2}{c}{UCF-QNRF} \\ \cline{3-10}
        && MAE & MSE & MAE & MSE & MAE & MSE & MAE & MSE \\
        \midrule[0.5pt]
        CAN \cite{liu2019context} & CVPR'19 & 62.3 & 100.0 & 7.8 & 12.2 & 212.2 & \textbf{243.7} & 107.0 & 183.0 \\
        Bayesian+ \cite{ma2019bayesian} & ICCV'19 & 62.8 & 101.8 & 7.7 & 12.7 & 229.3 & 308.2 & 88.7 & 154.8 \\ 
        S-DCNet \cite{xiong2019open} & ICCV'19 & 58.3 & 95.0 & 6.7 & 10.7 & 204.2 & 301.3 & 104.4 & 176.1 \\ 
        SANet + SPANet \cite{cheng2019learning} & ICCV'19 & 59.4 & 92.5 & 6.5 & \textbf{9.9} & 232.6 & 311.7 & - & - \\ 
        SDANet \cite{miao2020shallow} & AAAI'20 & 63.6&101.8 & 7.8&10.2 & 227.6&316.4 & -&- \\
        ADSCNet \cite{bai2020adaptive} & CVPR'20 & \underline{55.4} & 97.7 & \underline{6.4} & 11.3 & 198.4 & 267.3 & \textbf{71.3} & \textbf{132.5} \\ 
        ASNet \cite{jiang2020attention}  & CVPR'20 & 57.78 & \underline{90.13} & - & - & \underline{174.84} & \underline{251.63} & 91.59 & 159.71 \\  
        AMRNet \cite{liu2020adaptive} & ECCV'20 & 61.59&98.36 & 7.02&11.00 & 184.0&265.8 & 86.6&152.2 \\
        AMSNet \cite{hu2020count} & ECCV'20 & 56.7&93.4 & 6.7&10.2 & 208.4&297.3 & 101.8&163.2\\
        DM-Count \cite{wang2020distribution} & NeurIPS'20 & 59.7&95.7 & 7.4&11.8 & 211.0&291.5 & 85.6&\underline{148.3}\\
        Ours &- & \textbf{52.74} & \textbf{85.06} & \textbf{6.25} & \textbf{9.9} & \textbf{172.72} & 256.18 & \underline{85.32} & 154.5 \\  
        \bottomrule[1pt]
    \end{tabular}}
    }\vspace{0.2em}
    \caption{Comparison of the counting accuracy with state-of-the-art methods. }
    \label{tab:dataset}
    \vspace{-2.0em}
\end{table*}

As a comprehensive criteria, the proposed nAP metric is firstly reported to evaluate the performance of our P2PNet model. As shown in Table \ref{tab:loc}, the nAP is reported using three different thresholds of $\delta$, which corresponds to the average precision under different localization accuracies of the predicted individual points. Typically, nAP$_{0.5}$ could satisfy the requirements of most practical applications, which means that the ground truth point is exactly the nearest neighbor for most points within this region. Besides, nAP$_{0.1}$ and nAP$_{0.25}$ are reported to account for some requirements of high localization accuracy. Following recent detection methods which report the average of AP under several thresholds to provide a single number for the overall performance, we adopt a similar metric. Specifically, we calculate multiple nAP$_\delta$ with the $\delta$ starting from 0.05 to 0.50, with steps of 0.05. Then an average is done to get the overall average precision nAP$_{\{0.05:0.05:0.50\}}$.

From the Table \ref{tab:loc}, we observe that our P2PNet achieves a promising average precision under different levels of localization accuracy. Specifically, its overall metric nAP$_{\{0.05:0.05:0.50\}}$ is around 60\% on all datassets, which should already meet the requirements of many practical applications. In terms of the primary indicator nAP$_{0.5}$, the P2PNet generally achieves a promising precision of more than 80\%. For most datasets, the P2PNet could achieve a nAP$_{0.5}$ of nearly 90\%, which demonstrates the effectiveness of our approach on individual localization. Even for the \ywu{stricter} metric nAP$_{0.25}$, the precision is still higher than 55\%. These results are encouraging, since we did not use any techniques like coordinate refinement in \cite{cai2018cascade,zhang2018single} or exploiting multiple feature levels \cite{lin2017feature}, which are both orthogonal to our contributions and should bring more improvements. Besides, the P2PNet achieves a relatively lower precision on the nAP$_{0.05}$, which is reasonable since the effects of the labeling deviations might gradually become apparent under such high localization accuracy.

Besides, we also notice that the NWPU-Crowd dataset \cite{wang2020nwpu} provides scarce yet valuable box annotations, so we report our localization performance using their metrics to compare with other competitors. And our P2PNet achieves an F1-measure/Precision/Recall of 71.2\%/72.9\%/69.5\%, which is the best among published methods with similar backbones. For other localization based methods with official codes available, we also report their results in nAP metric (much lower than ours) in \textbf{Supplementary}.

Furthermore, we also evaluate the counting accuracy of our model. The estimated crowd number of our P2PNet is obtained by counting the predicted points with confidence scores higher than 0.5. We compare the P2PNet with state-of-the-art methods on several challenging datasets with various densities. Similar to \cite{zhang2016single}, we also adopt Mean Absolute Error (MAE) and Mean Squared Error (MSE) as the evaluation metrics. The results are illustrated in Table \ref{tab:dataset} and Table \ref{tab:nwpu_data}. The top performance is indicated by bold numbers and the second best is indicated by underlined numbers.
\begin{table}[ht!]
\centering
\small{
\begin{tabular}{c|cccc}
\toprule[1pt]
\multirow{2}{*}{Methods}&\multicolumn{4}{c}{NWPU-Crowd}\\ \cline{2-5}
& MAE[O]&MSE[O]&MAE[L]&MAE[S] \\ \hline
CSRNet \cite{li2018csrnet} & 121.3 & 387.8 & 112.0 & \underline{522.7} \\
Bayesian+ \cite{ma2019bayesian} & 105.4 & 454.2 & 115.8 & 750.5 \\
S-DCNet \cite{xiong2019open}  & 90.2 & 370.5 & \textbf{82.9} & 567.8 \\
DM-Count \cite{wang2020distribution} & \underline{88.4} & 388.6 & 88.0 & \textbf{498.0} \\
\textbf{Ours} & \textbf{77.44}&\textbf{362} & \underline{83.28}& 553.92 \\
\bottomrule[1pt]
\end{tabular}}
\vspace{-0.5em}
\caption{Comparison on the NWPU-Crowd dataset.}\label{tab:nwpu_data}
\vspace{-1.5em}
\end{table}
\begin{figure*}[t!] 
\vspace{-1.0em}
  \centering
  \includegraphics[width=0.96\textwidth]{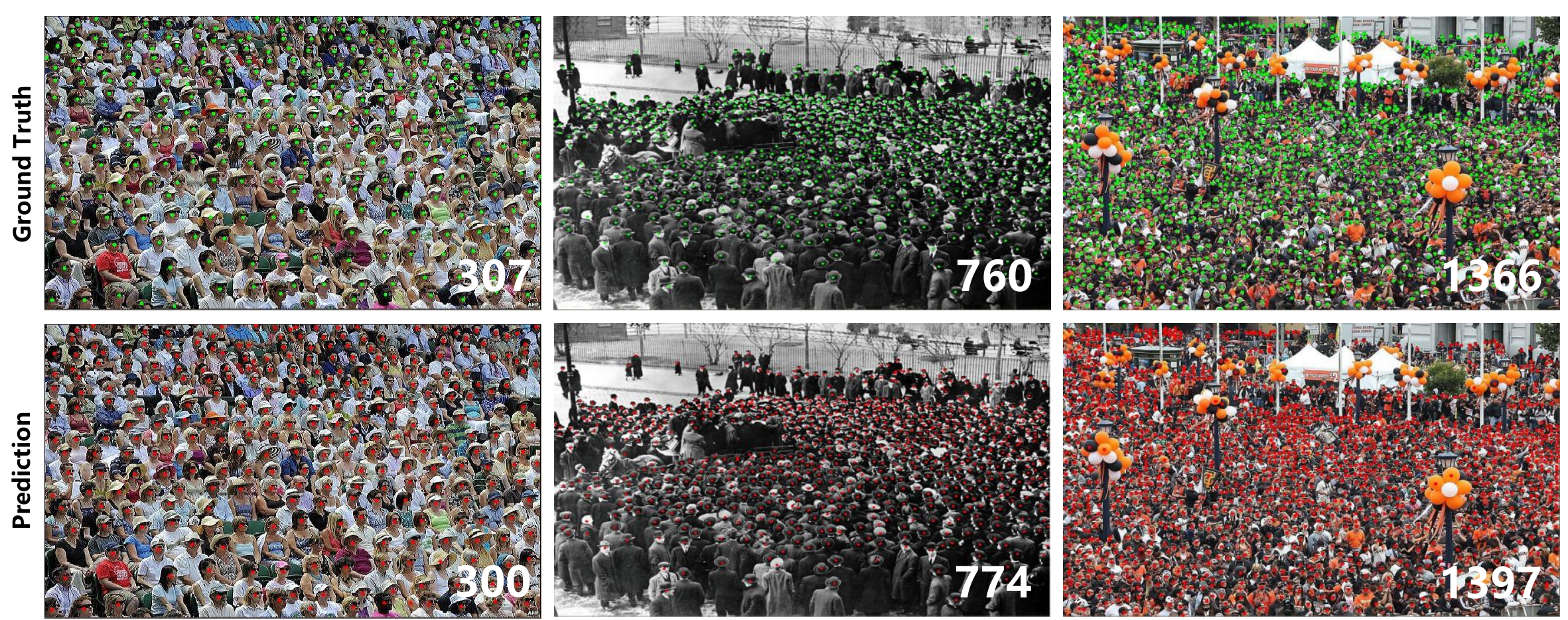}
  \caption{Some qualitative results for the predicted individuals of our P2PNet. The white numbers denote to the ground truth or prediction counts. The visualizations demonstrate the superiority of our method under various densities in terms of both localization and counting.}
  \label{fig6} 
  \vspace{-1.5em}
\end{figure*}

\noindent\textbf{ShanghaiTech}. There are two independent subsets in ShanghaiTech dataset: PartA and PartB. The PartA contains highly congested images collecting from the Internet. While the PartB is collected from a busy street and represents relatively sparse scenes. Our P2PNet achieves the best performance on both PartA and PartB. In particular, on the PartA, the P2PNet reduces the MAE by 4.8\% and MSE by 12.9\% respectively, compared with the second best method ADSCNet. For sparse scenes in PartB, the P2PNet could also bring a reduction of 2.3\% in MAE. \\
\noindent\textbf{UCF\_CC\_50}. UCF\_CC\_50 has only 50 images collecting from the Internet, but contains complicated scenes with large variation of crowd numbers. As shown in Table \ref{tab:dataset}, our P2PNet surpasses all the other methods, reducing the MAE by 2.1 compared with the second best performance. \\
\noindent\textbf{UCF-QNRF}. UCF-QNRF is a challenging dataset due to the much wider range of counts. As shown in Table \ref{tab:dataset}, our P2PNet achieves an MAE of 85.32, which is much better than the Neural Architecture Search based method AMSNet. Compared with the previous best method ADSCNet, although the accuracy of our method is not so competitive, it is still much higher than ADSCNet on all other datasets. Besides, among all the methods in Table \ref{tab:dataset}, only \ywu{ours} could provide \ywu{exact individual locations}.

\noindent\textbf{NWPU-Crowd}. The NWPU-Crowd dataset is a large-scale congested dataset recently introduced in \cite{wang2020nwpu}. As shown in Table \ref{tab:nwpu_data}, our P2PNet achieves the best overall MAE, with a reduction of 12.4\% compared with the second best method DM-Count. Since our predictions are only based on a single scale feature map for simplicity, the result is slightly lower than those best performance on MAE[S]. MAE[S] is the average MAE of different scale levels, please refer to \cite{wang2020nwpu}. 

\subsection{Ablation Studies}
\begin{table}[!ht]
\vspace{-1.0em}
  \centering
  \begin{tabular}{c|c c c}
    \toprule[1pt]
    Layout & MAE & MSE & nAP$_{\delta}$\\
    \hline
    Center &  53.7 & 89.61 & 61.7\\
    \hline
    Grid &   \textbf{52.74} &  \textbf{85.06}  & \textbf{64.4}  \\
    \bottomrule[1pt]
  \end{tabular}
  \caption{The effect of the layout for reference points. For an overall comparison, we use $\delta=\{{0.05:0.05:0.50}\}$.}\label{tab:anchor}
    \vspace{-1.5em}
\end{table}
\noindent \textbf{Layout of reference points.} We firstly evaluate the effect from the layouts of the reference points. As shown in Table \ref{tab:anchor}, we compare two layouts in the Figure \ref{fig4}. Generally speaking, both the two layouts achieve state-of-the-art performance with minor difference, proving that the target association matters more than the layout of reference points. The Grid layout performs slightly better due to its dense arrangement of reference points, which is beneficial for the congested regions.
\begin{table}[!ht]
  \centering
  \vspace{-0.5em}
  \begin{tabular}{c|c|c c c}
    \toprule[1pt]
    \multicolumn{2}{c|}{Method} & MAE & MSE & nAP$_{\delta}$\\
    \hline
    \multirow{3}{*}{P2PNet} & $s=4$ & 53.51 & 85.77 & \textbf{66.8} \\
                               & $s=8$ & \textbf{52.74} &  \textbf{85.06} & 64.4\\
                               & $s=16$ & 54.3  &  85.18 & 52.4 \\
    \bottomrule[1pt]
  \end{tabular}
  \caption{The ablation study on SHTech PartA. For an overall comparison, we use $\delta=\{{0.05:0.05:0.50}\}$.}\label{tab:strides}
    \vspace{-1.5em}
\end{table}

\noindent \textbf{Effect of feature levels.} We exhibit the effect of different feature levels used for prediction. For fair comparison, we keep the total reference points the same when using feature levels with different strides. As shown in Table \ref{tab:strides}, the P2PNet consistently achieves competitive results using different feature levels, which demonstrates the effectiveness of our point based solution. In particular, the feature level with a stride of 8 provides a trade-off for the various densities, thus yields better performance.

In terms of the localization accuracy, we observe an obvious trend of improvement on nAP when we increase the feature map resolution, as shown in Table \ref{tab:strides}. \ywu{It} implies that the finest feature map is beneficial for \ywu{localization}, which is also in accord with the consensus on other tasks. Besides, based on our baseline method, it would be interesting to introduce existing multi-scale feature fusion techniques such as \cite{lin2017feature}, which are discarded in our P2PNet for simplicity.
\vspace{-0.6em}
\section{Conclusion}
\vspace{-0.6em}
In this work, we go beyond crowd counting and propose a purely point-based framework to directly \ywu{predict locations for crowd individuals}. This new framework could better satisfy the practical demands of downstream tasks in crowd analysis. \ywu{In} conjunction with \ywu{it}, we advocate to use \ywu{a} new metric nAP for a more comprehensive \ywu{accuracy} evaluation on both localization and counting. \ywu{Moreover}, as an intuitive solution following this framework, we propose \ywu{a novel network} P2PNet, which is capable of directly \ywu{taking point annotations as supervision whilst} predicting the point locations \ywu{during inference}. \ywu{P2PNet's} key component is the one-to-one matching during the ground truth targets association, which is beneficial to the improvement of the nAP metric. This conceptually simple framework yields state-of-the-art counting performance and promising localization accuracy. 

{\small
\bibliographystyle{ieee_fullname}
\bibliography{arXiv}
}

\newpage

\begin{center}
\textbf{{\Large Supplementary}}
\end{center}
\setcounter{section}{0}
\section{Counting Evaluation Metrics}\label{metric}
Similar to previous works in crowding counting, we adopt Mean Absolute Error (MAE) and Mean squared error (MSE) as our evaluation metrics which are defined as:
\begin{equation}
    \mathrm{MAE} = \frac{1}{N} \sum_i^{N} \left|\hat{z}_{i} - z_{i}\right|,
\end{equation}
\begin{equation}
    \mathrm{MSE} = \sqrt{\frac{1}{N} \sum_i^{N} (\hat{z}_{i} - z_{i})^2},
\end{equation}
where $\hat{z}_{i}$ and $z_{i}$ represent estimated crowd number and ground-truth crowd number of the $i$-th image\ywu{,} respectively. $N$ denotes the total number of \ywu{test} images.
\section{Discussion on Spatial Scale Problem}\label{dis}
Despite its superior performance, the proposed P2PNet did not explicitly deal with the scale variation problem. Actually, different from bounding boxes, the head points themselves are scale ignorant in nature. In other words, the one-one matching ensures that no matter which scale the head is, only one optimal predicted proposal will be chosen as its prediction. Thus, some implicit scale cues might be learned automatically during the training process. Besides, the proposed framework is orthogonal to some previous works dealing with scale variations, such as FPN \cite{lin2017feature}, PGCNet \cite{pgc2019}, CSRNet \cite{li2018csrnet}, MCNN \cite{zhang2016single}, \textit{etc}.

\section{Hyperparameters Analysis}\label{ana}
We set the number \ywu{of} reference points \ywu{($K$)} based on the nearest neighbour distance distribution of ground truth points. Specifically, based on the observation that nearly 95\% (SHTech PartA) of the head points are \ywu{within} the nearest neighbour distance of 4 pixels, we set the number of the reference points $K$ as 4 on the feature map with stride 8. We experimentally analyze the accuracy sensitivity of this parameter in Table \ref{tab:num}. As shown from the results, the model with $K$=1 still achieves state-of-the-art accuracy, although it's reference points are too \ywu{few} to cover all the heads in congested areas. Setting $K$ to a value greater than 4 leads to inferior accuracy, which might be caused by the \ywu{increase} of negative samples.

\begin{table}[!ht]
\setlength\tabcolsep{18pt}
  \centering
  \begin{tabular}{c|c c c}
    \toprule[1pt]
    $K$ & MAE & MSE & nAP$_{\delta}$\\
    \hline
    $K$=1 &  54.08  &  \textbf{84.37} & 60.1 \\
    $K$=4 &  \textbf{52.74}  &  85.06 & \textbf{64.4} \\
    $K$=8 &   53.43 &  87.57 &  58.8\\
    $K$=12 &   54.13 &  87.9 & 58.6 \\
    $K$=16 &  53.47  & 86.1  & 58.3  \\
    \bottomrule[1pt]
  \end{tabular}
  \caption{The \ywu{performance change w.r.t.} the number $K$ for reference points. For an overall comparison, we use $\delta=\{{0.05:0.05:0.50}\}$.}\label{tab:num}
\end{table}

\section{Localization Performance}\label{loc}
\begin{table}[!ht]
\setlength\tabcolsep{8pt}
  \centering
  \begin{tabular}{c|c c c}
    \toprule[1pt]
    Method & F1-Measure & Precision & Recall\\
    \hline
    FasterRCNN \cite{ren2016faster} &  0.068 &  0.958 & 0.035 \\
    TinyFaces \cite{hu2017finding}&  0.567  &  0.529 & 0.611 \\
    RAZ \cite{liu2019recurrent}&   0.599 &  0.666 &  0.543\\
    Crowd-SDNet \cite{wang2021self} &  0.637  & 0.651  & 0.624  \\
    PDRNet \cite{li2019pdr}&  0.653 & 0.675  & 0.633  \\
    TopoCount \cite{abousamra2020localization} & 0.692  & 0.683  & \textbf{0.701} \\
    D2CNet \cite{cheng2021decoupled} & \underline{0.700} & \textbf{0.741}  & 0.662 \\
    Ours &\textbf{0.712} & \underline{0.729}  & \underline{0.695} \\
    \bottomrule[1pt]
  \end{tabular}
  \caption{Comparison for the localization performance on NWPU.}\label{tab:f1}
\end{table}
Thanks to the scarce yet valuable box annotations provided by the NWPU-Crowd dataset \cite{wang2020nwpu}, we could compare the localization performance of our P2PNet with other competitors using their metrics. As shown in Table \ref{tab:f1}, our P2PNet achieves the best F1 score among the published methods with similar computation complexity.

Among a few existing localization-based methods, almost none of them have official codes
or third-party re-implementations except for \cite{sam2020locate}. So for a fair comparison, we evaluate the nAP$_{0:05:0:05:0:50}$ of \cite{sam2020locate} on SHTech
PartA, SHTech PartB and QNRF, which are 33.2\%, 45.8\% and 8.9\% respectively. As shown from the results, our P2PNet achieves significantly higher localization performance in terms of nAP, especially on the challenging QNRF dataset.

\section{\ywu{Visual Results for Qualitative Evaluation}}\label{vis}
\begin{figure*}[t!] 
\vspace{-1.0em}
  \centering
  \includegraphics[width=1.05\textwidth]{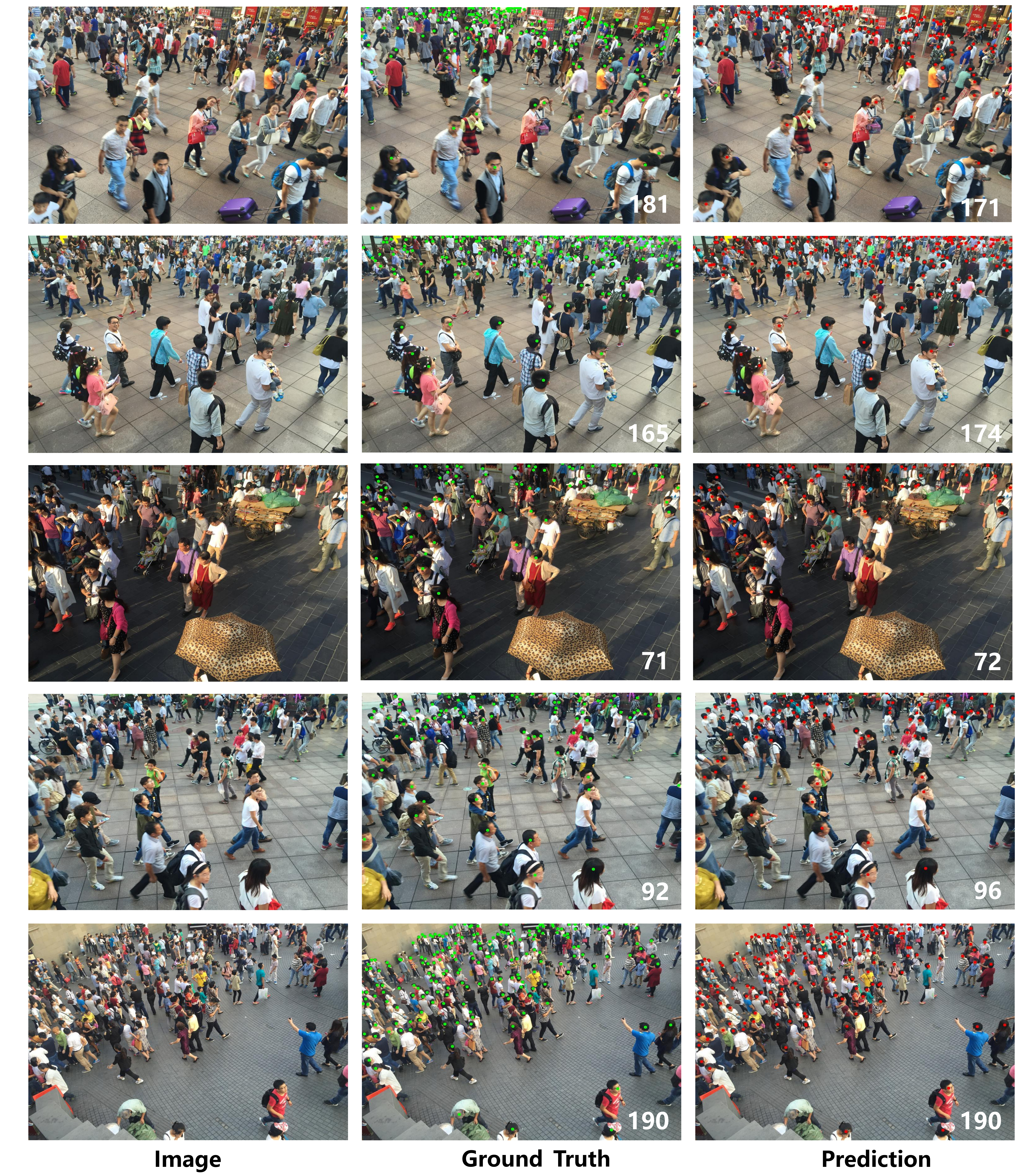}
  \caption{\ywu{Visual results} of sparse scenes (1).}
  \label{supp_fig1} 
  \vspace{-1.5em}
\end{figure*}

\begin{figure*}[t!] 
\vspace{-1.0em}
  \centering
  \includegraphics[width=1.05\textwidth]{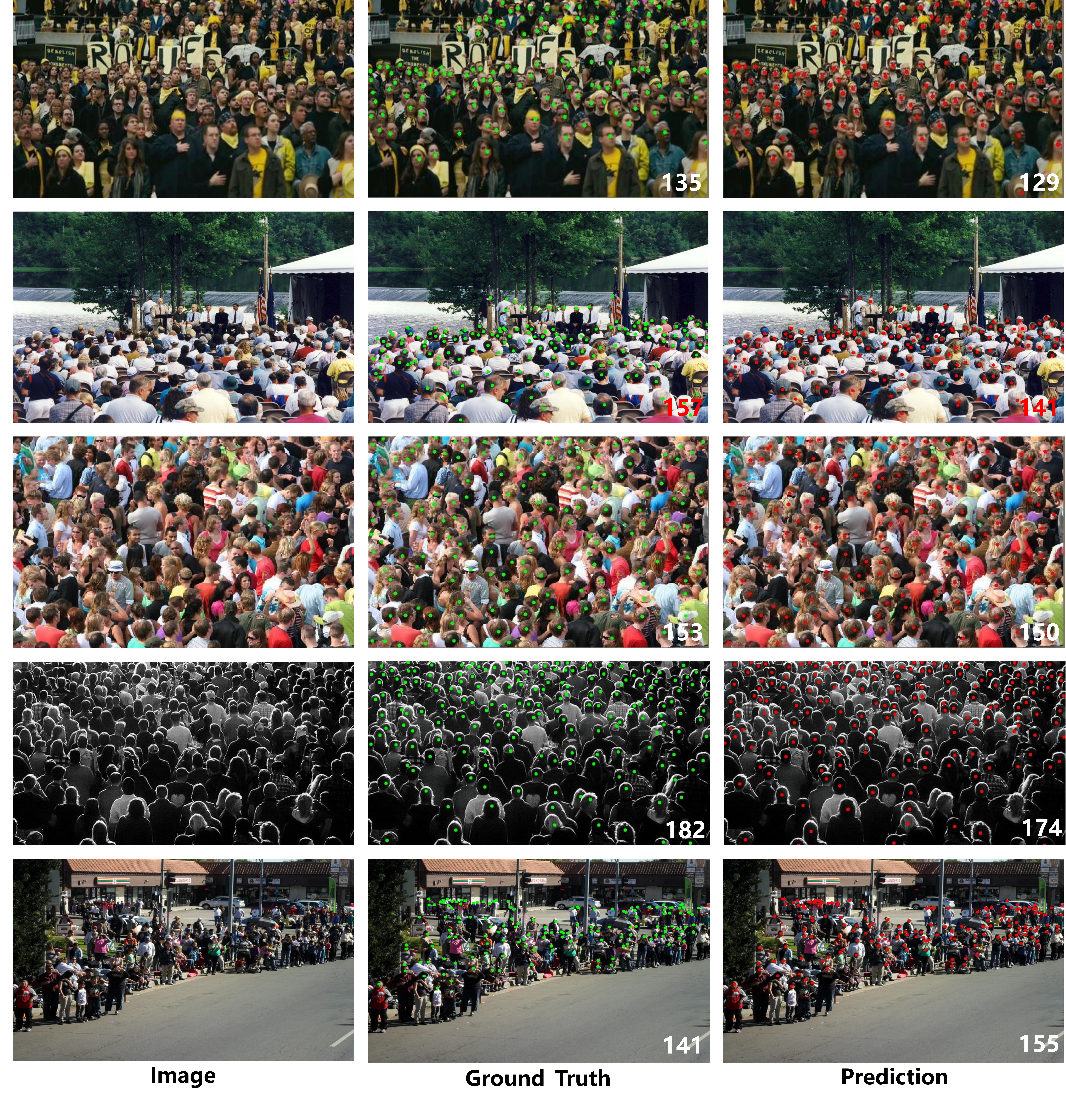}
  \caption{\ywu{Visual results} of sparse scenes (2).}
  \label{supp_fig2} 
  \vspace{-1.5em}
\end{figure*}

\begin{figure*}[t!] 
\vspace{-1.0em}
  \centering
  \includegraphics[width=1\textwidth]{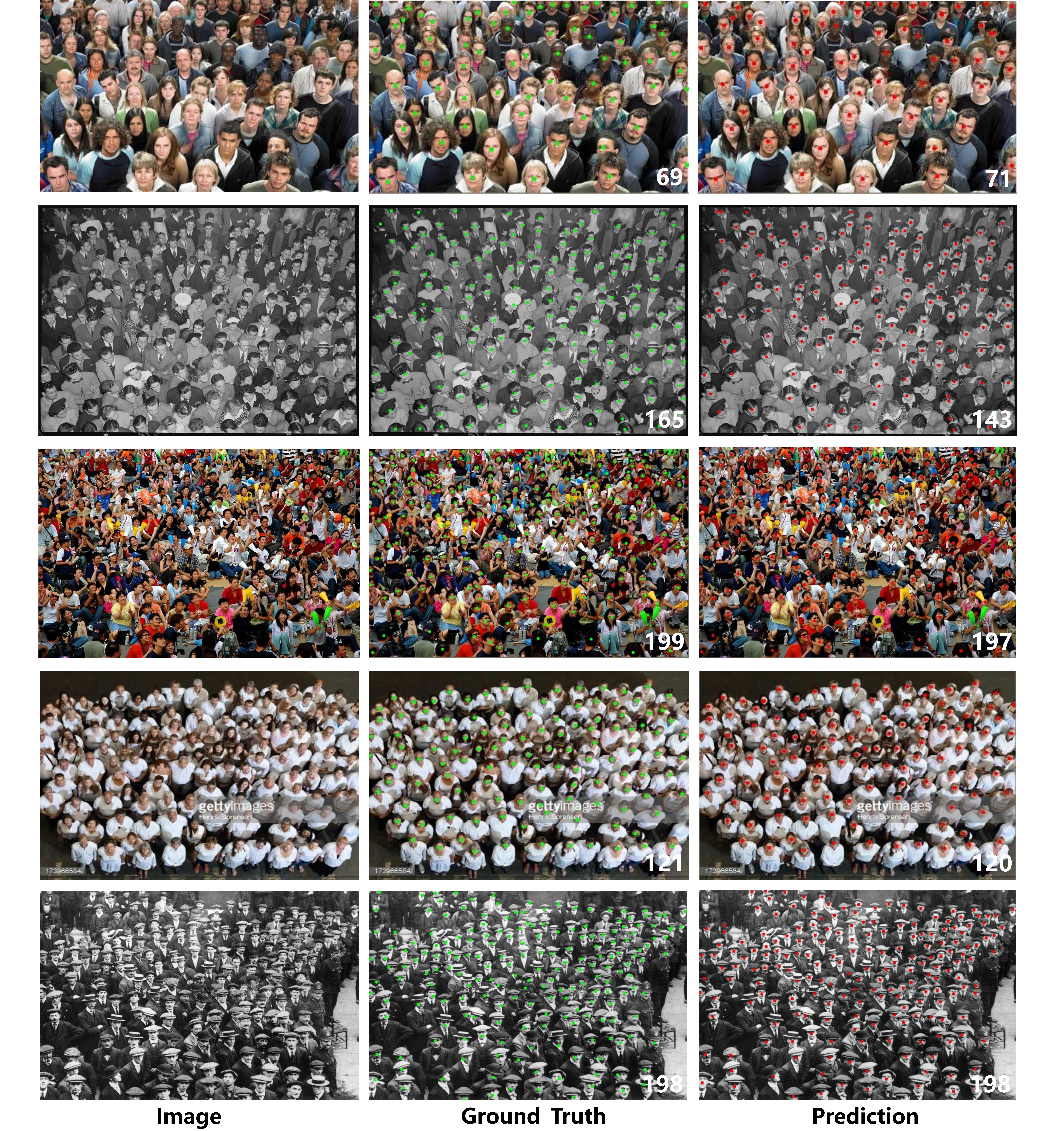}
  \caption{\ywu{Visual results} of sparse scenes (3).}
  \label{supp_fig3} 
  \vspace{-1.5em}
\end{figure*}

\begin{figure*}[t!] 
\vspace{-1.0em}
  \centering
  \includegraphics[width=1\textwidth]{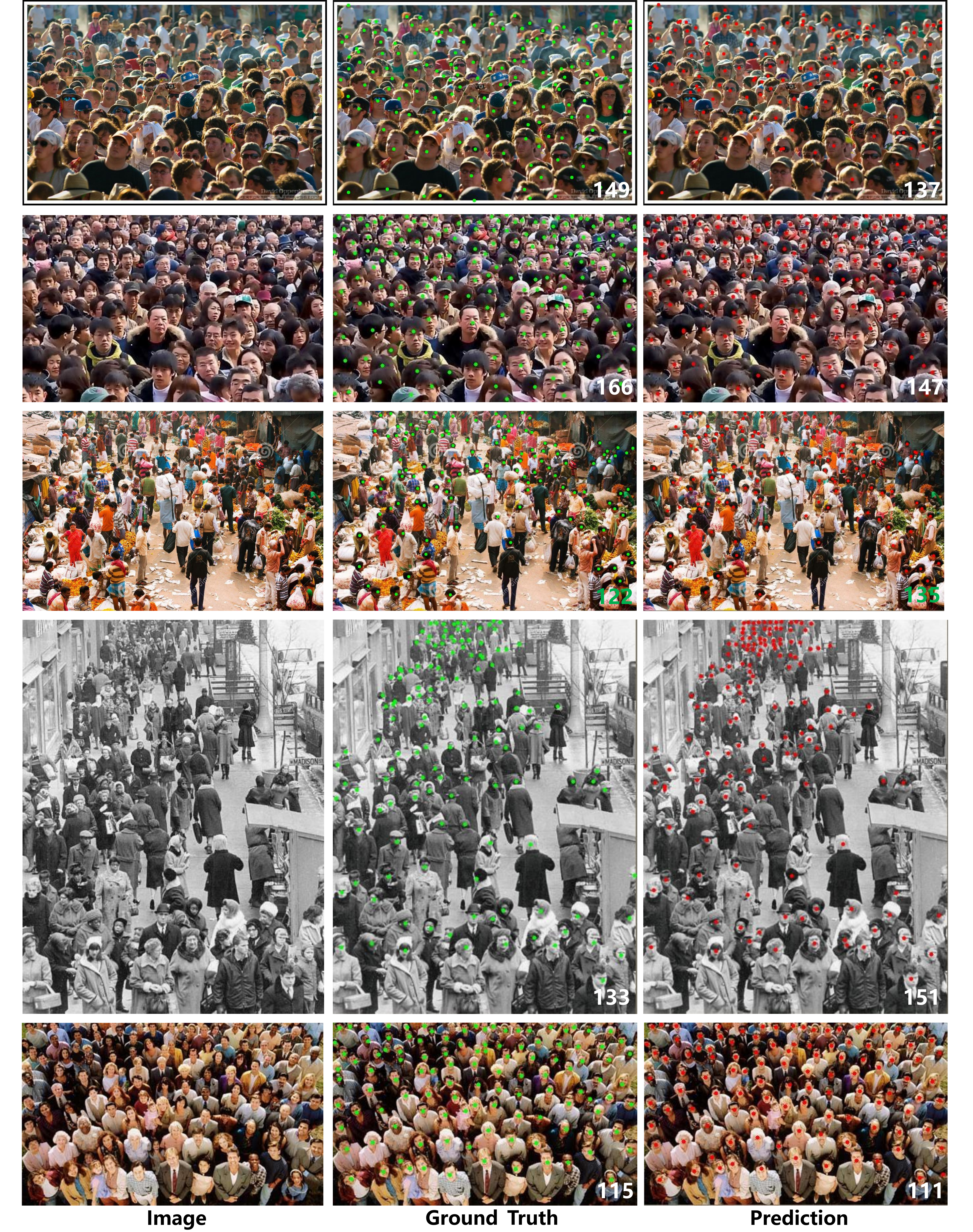}
  \caption{\ywu{Visual results} of sparse scenes (4).}
  \label{supp_fig4} 
  \vspace{-1.5em}
\end{figure*}

\begin{figure*}[t!] 
\vspace{-1.0em}
  \centering
  \includegraphics[width=1\textwidth]{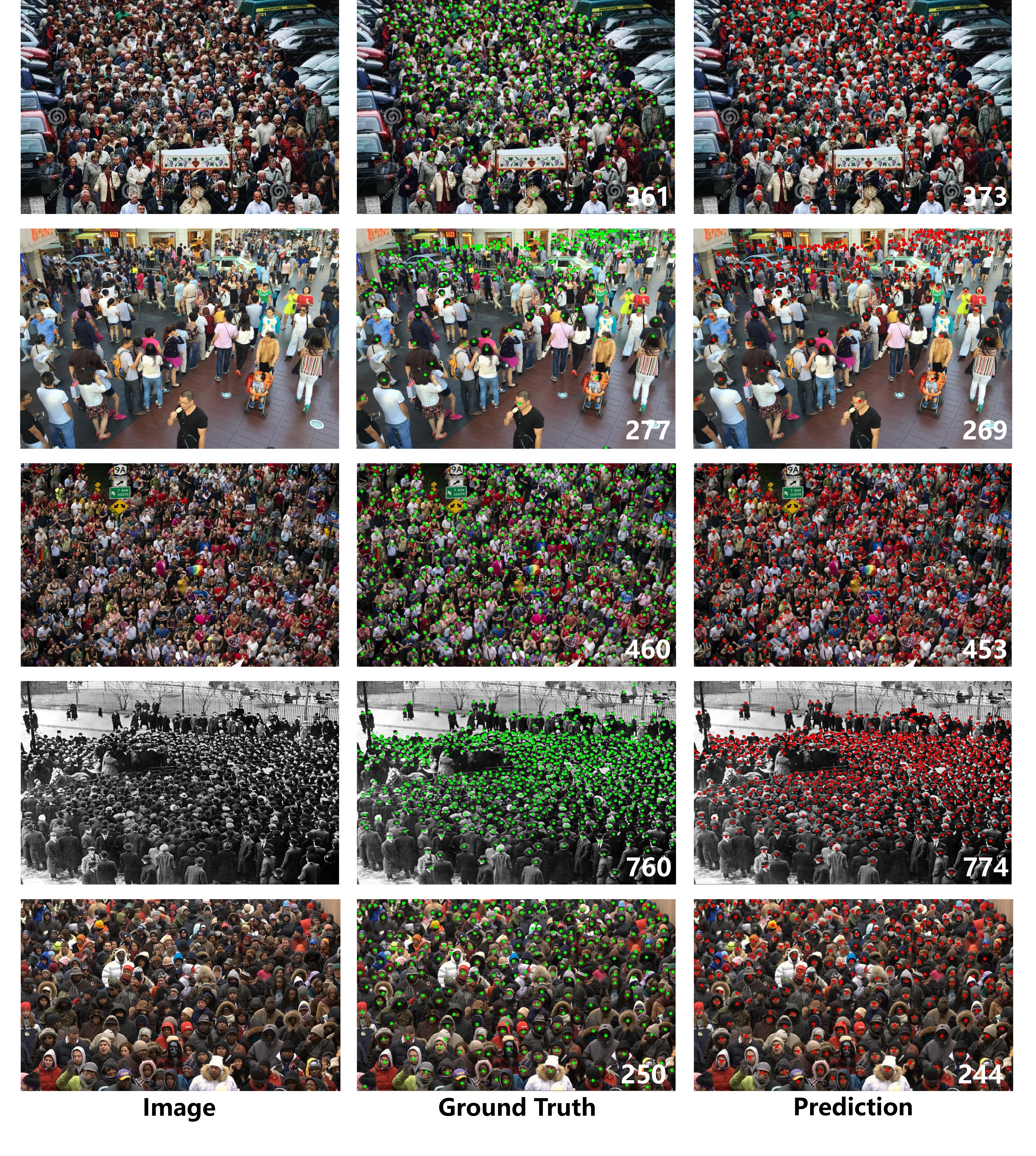}
  \caption{\ywu{Visual results} of moderately congested scenes (1).}
  \label{supp_fig5} 
  \vspace{-1.5em}
\end{figure*}

\begin{figure*}[t!] 
\vspace{-1.0em}
  \centering
  \includegraphics[width=1\textwidth]{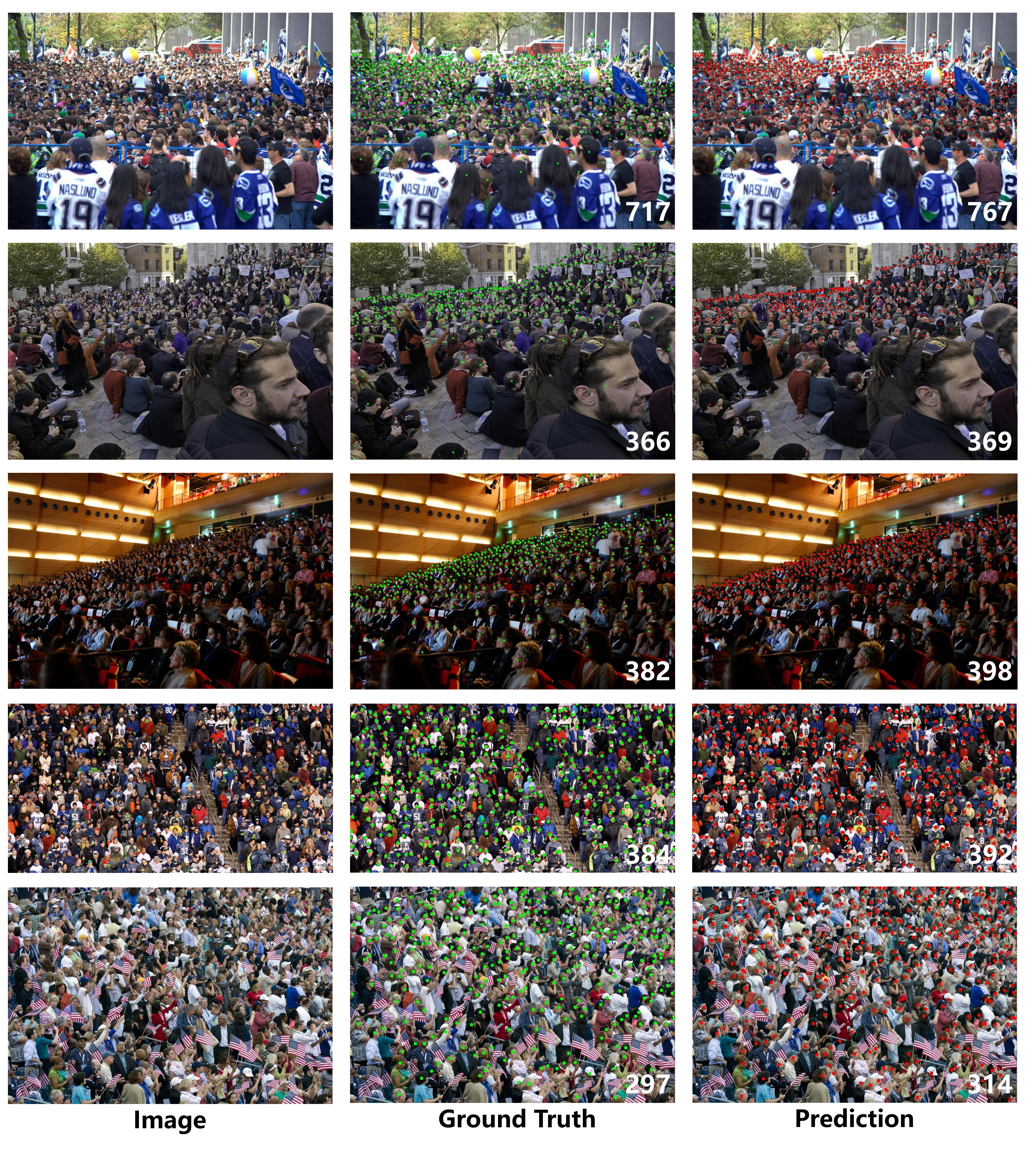}
  \caption{\ywu{Visual results} of moderately congested scenes (2).}
  \label{supp_fig6} 
  \vspace{-1.5em}
\end{figure*}

\begin{figure*}[t!] 
\vspace{-1.0em}
  \centering
  \includegraphics[width=1\textwidth]{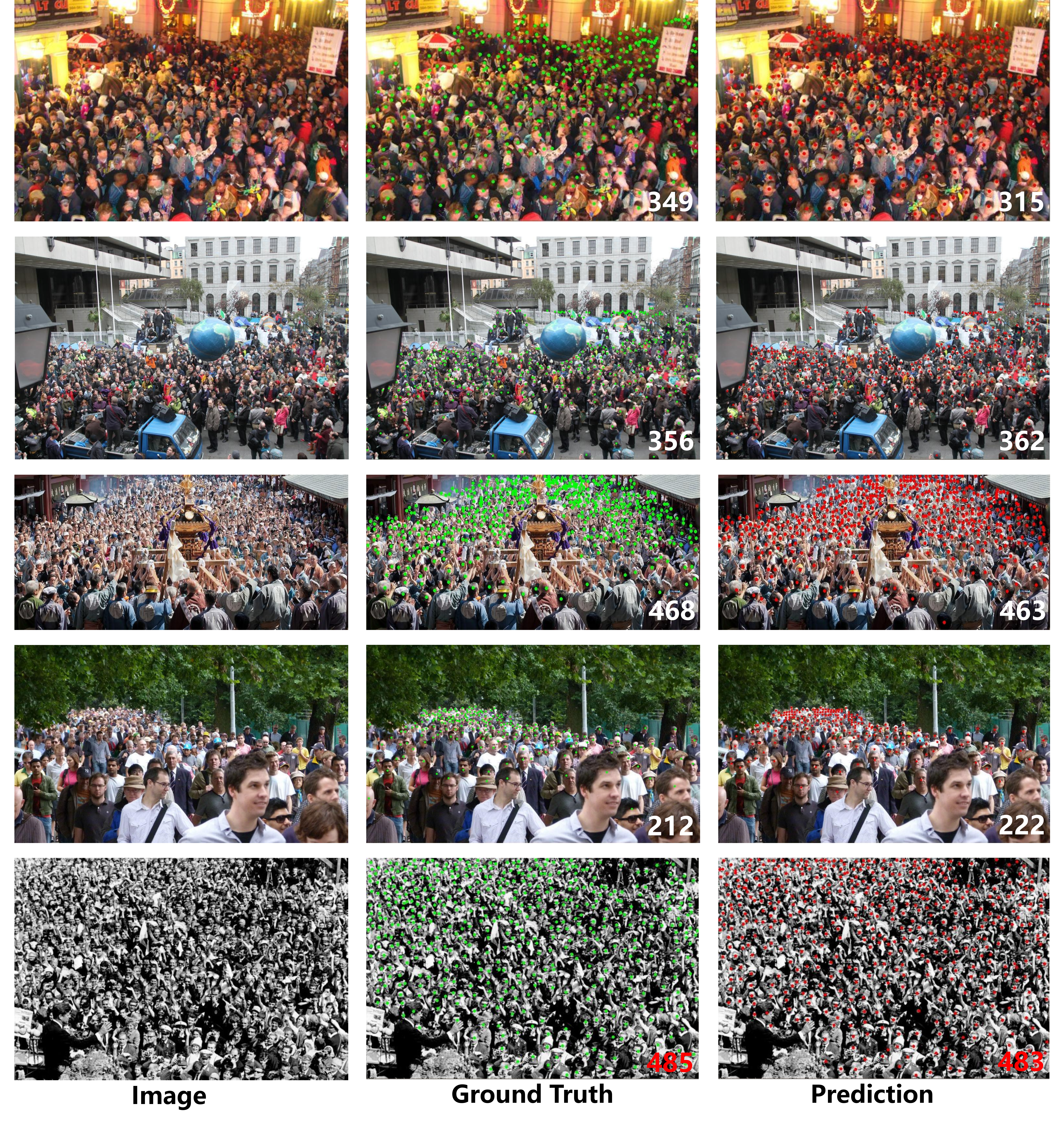}
  \caption{\ywu{Visual results} of moderately congested scenes (3).}
  \label{supp_fig7} 
  \vspace{-1.5em}
\end{figure*}

\begin{figure*}[t!] 
\vspace{-1.0em}
  \centering
  \includegraphics[width=1\textwidth]{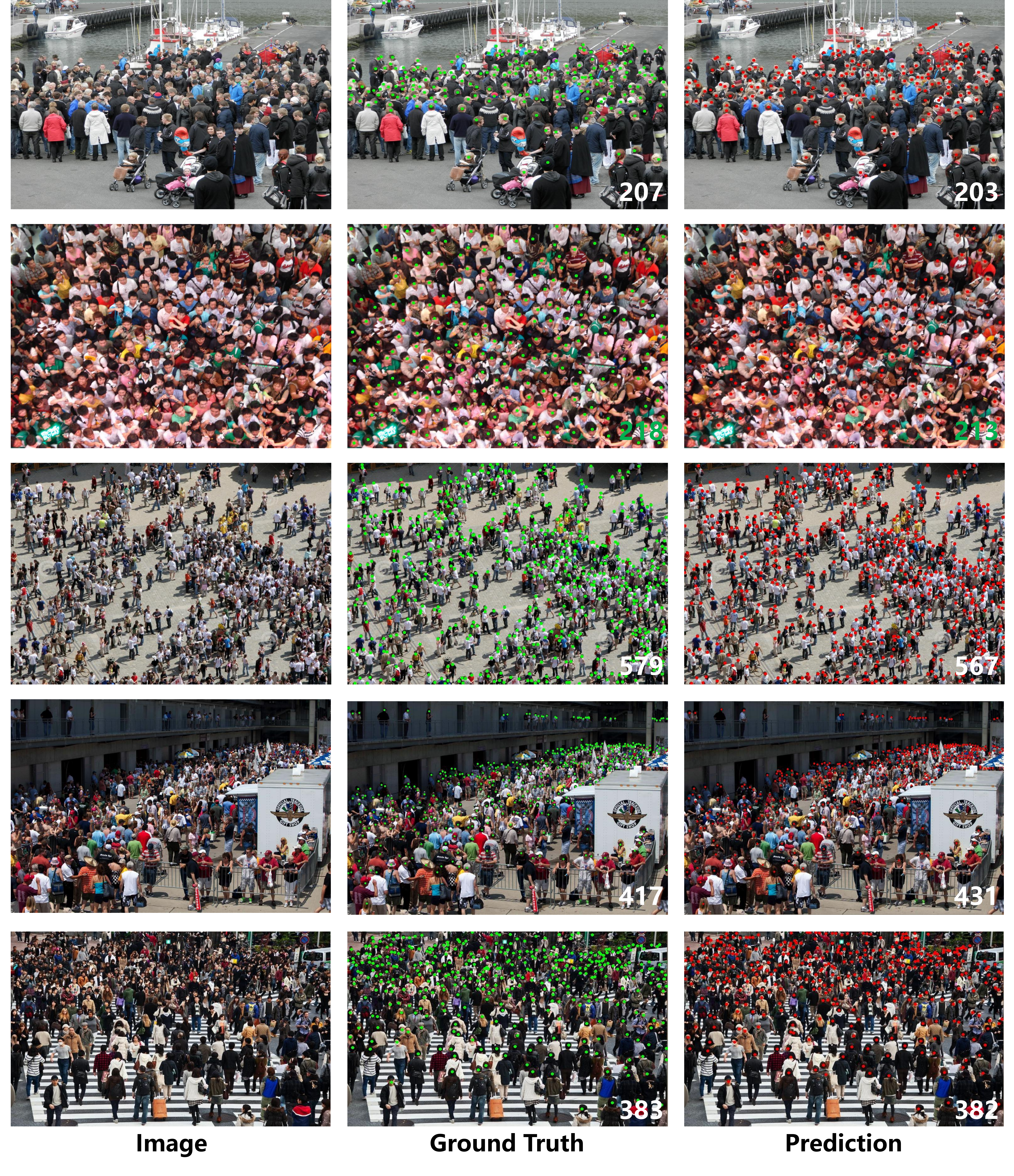}
  \caption{\ywu{Visual results} of moderately congested scenes (4).}
  \label{supp_fig8} 
  \vspace{-1.5em}
\end{figure*}

\begin{figure*}[t!] 
\vspace{-1.0em}
  \centering
  \includegraphics[width=1\textwidth]{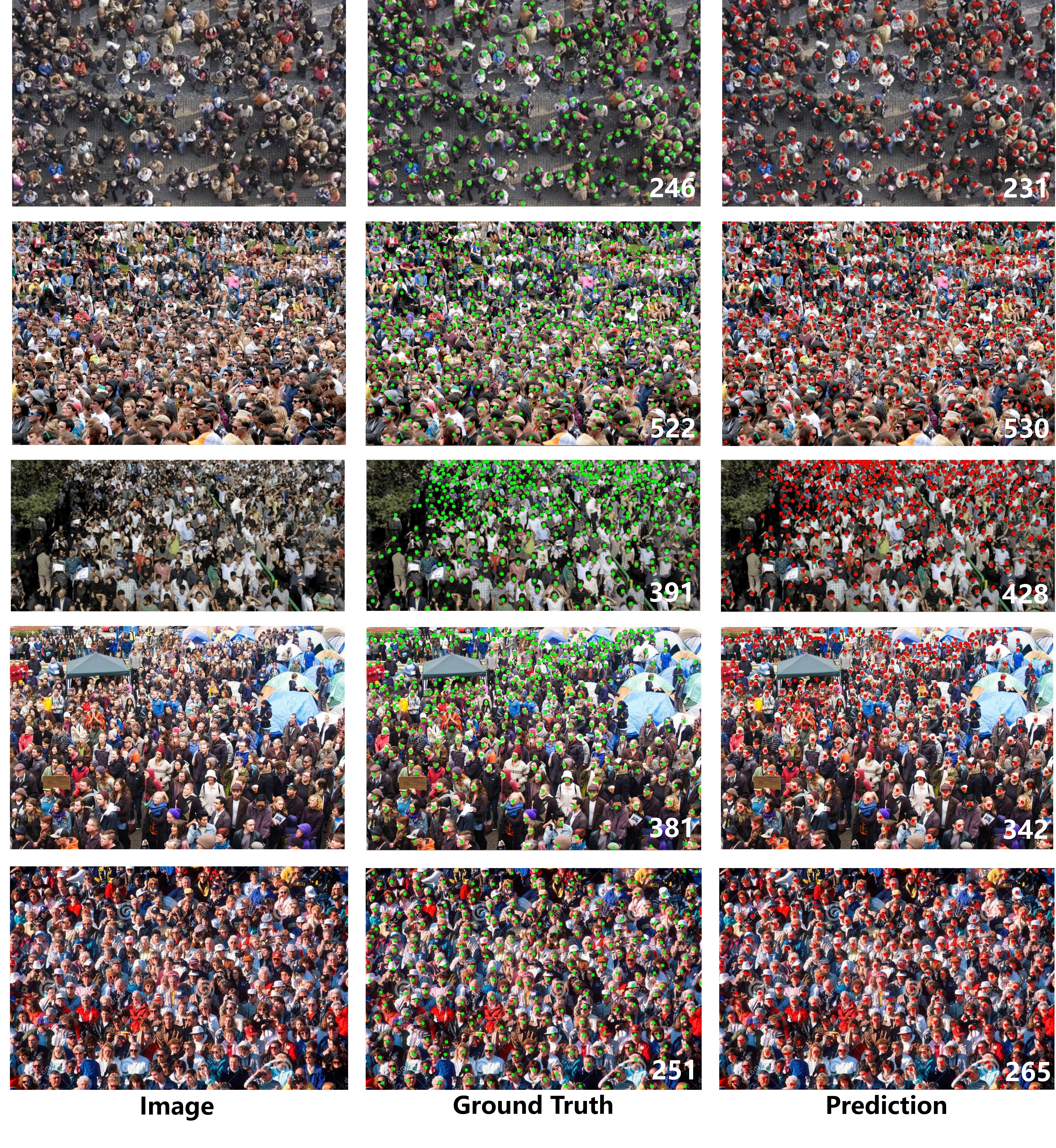}
  \caption{\ywu{Visual results} of moderately congested scenes (5).}
  \label{supp_fig9} 
  \vspace{-1.5em}
\end{figure*}

\begin{figure*}[t!] 
\vspace{-1.0em}
  \centering
  \includegraphics[width=1\textwidth]{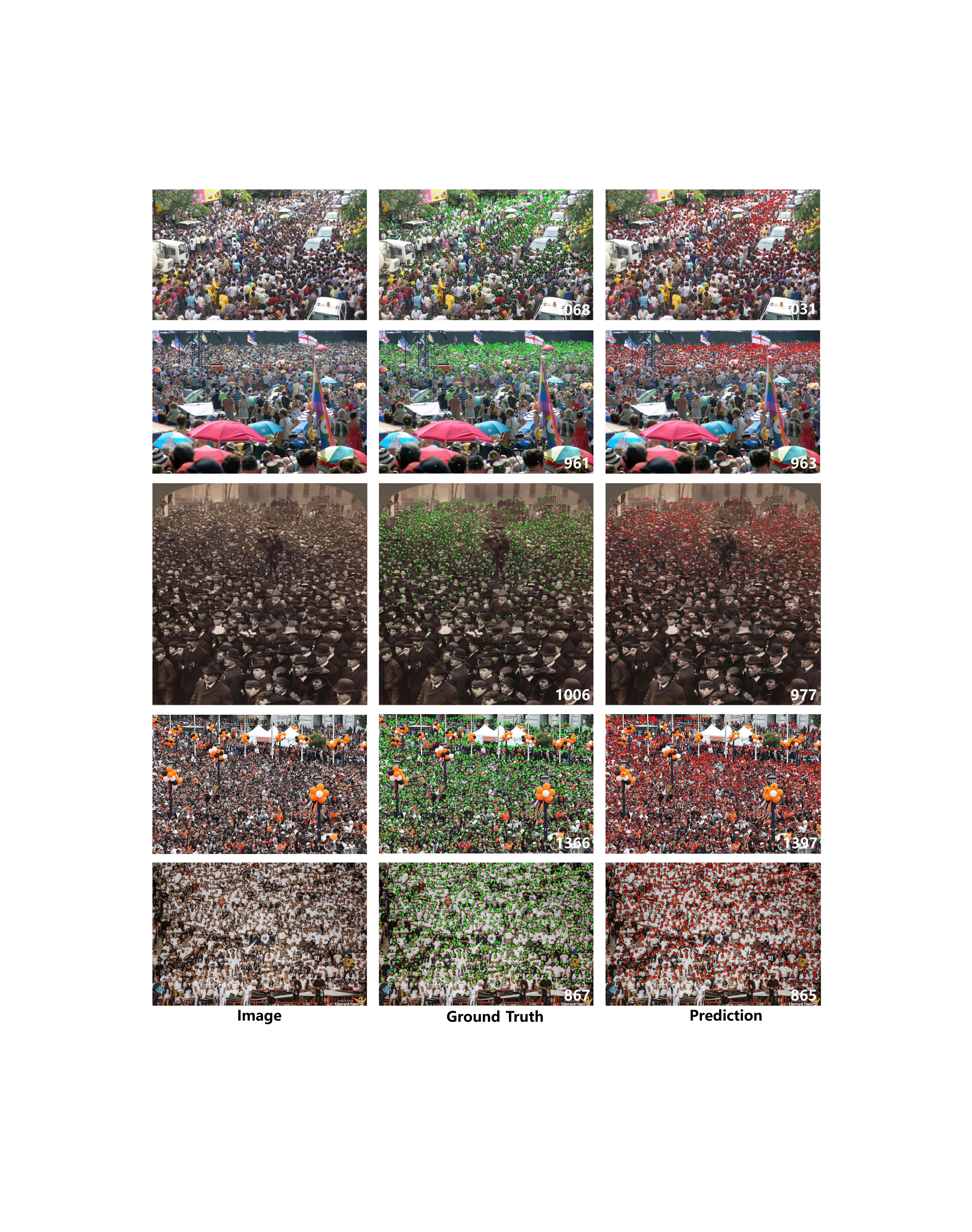}
  \caption{\ywu{Visual results} of congested scenes (1).}
  \label{supp_fig10} 
  \vspace{-1.5em}
\end{figure*}

\begin{figure*}[t!] 
\vspace{-1.0em}
  \centering
  \includegraphics[width=1\textwidth]{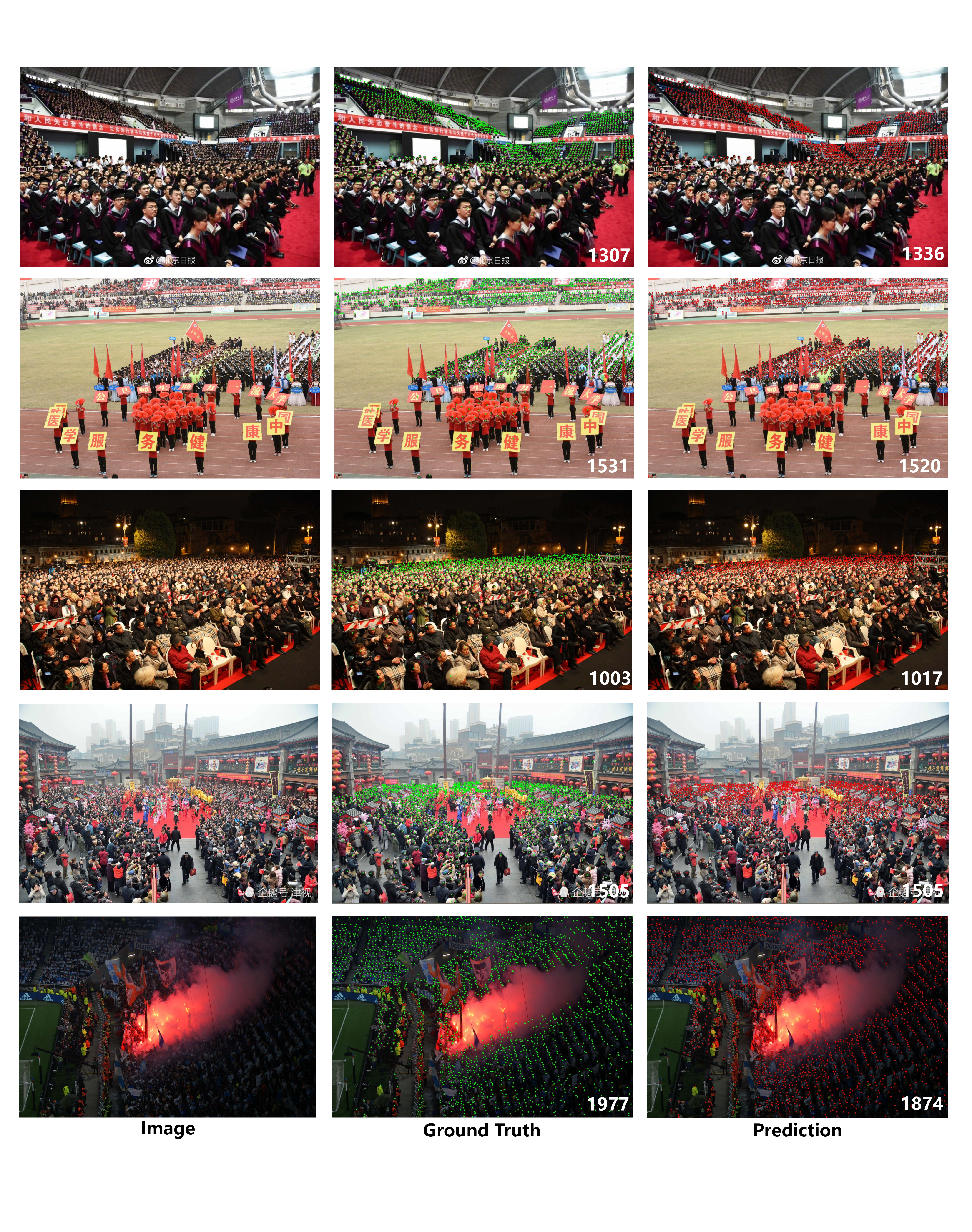}
  \caption{\ywu{Visual results} of congested scenes (2).}
  \label{supp_fig11} 
  \vspace{-1.5em}
\end{figure*}

\begin{figure*}[t!] 
\vspace{-1.0em}
  \centering
  \includegraphics[width=1\textwidth]{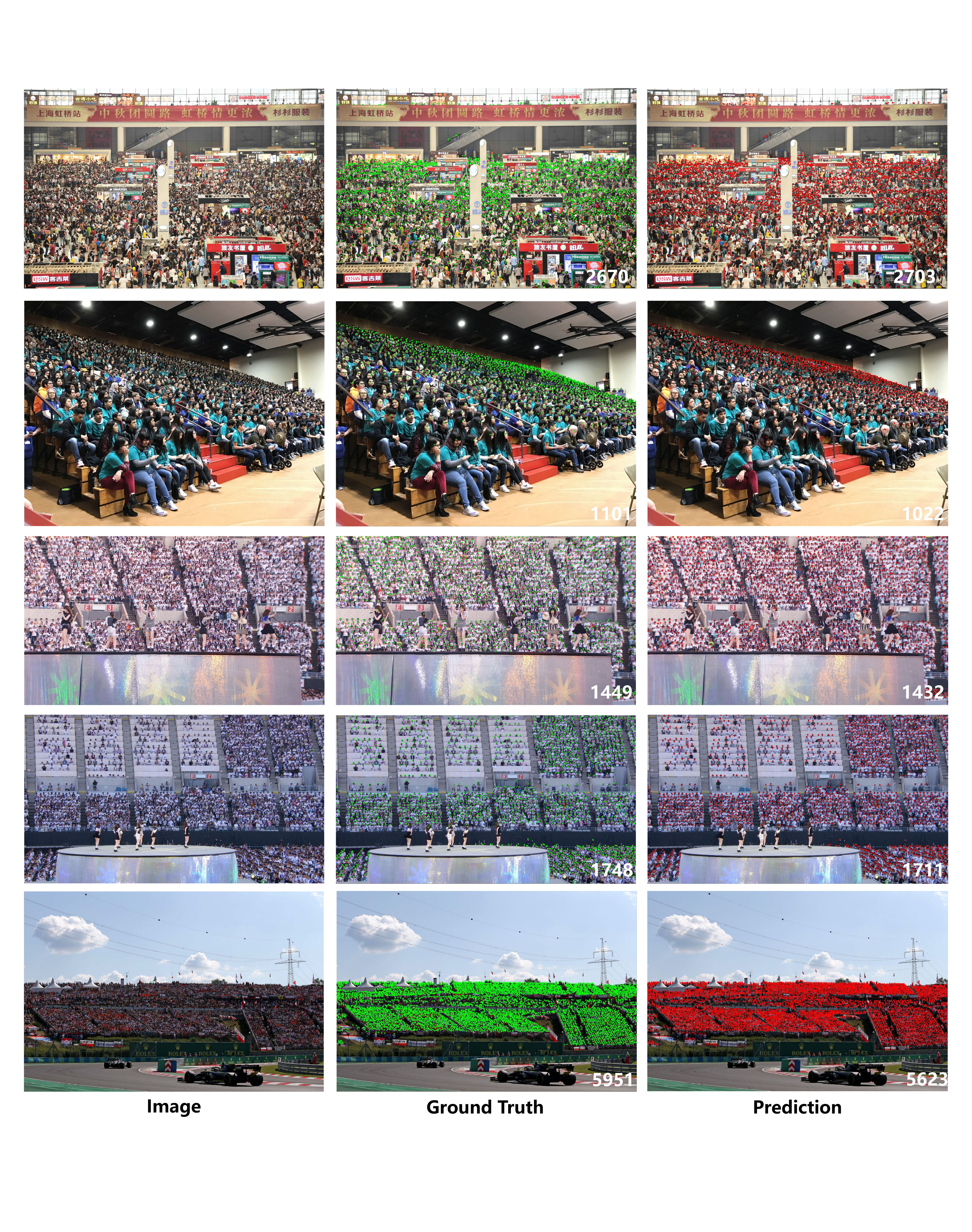}
  \caption{\ywu{Visual results} of congested scenes (3).}
  \label{supp_fig12} 
  \vspace{-1.5em}
\end{figure*}

\begin{figure*}[t!] 
\vspace{-1.0em}
  \centering
  \includegraphics[width=1\textwidth]{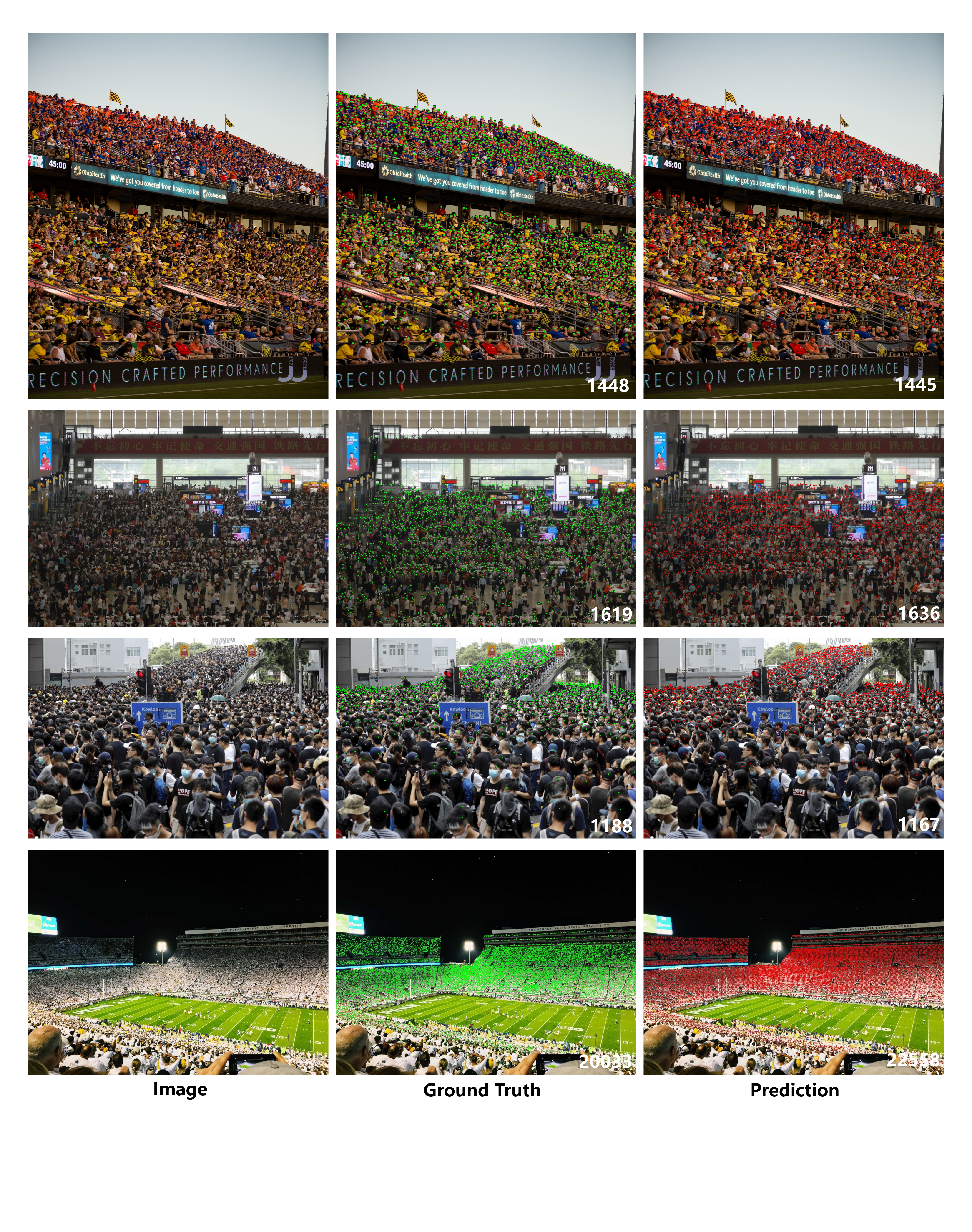}
  \caption{\ywu{Visual results} of congested scenes (4).}
  \label{supp_fig13} 
  \vspace{-1.5em}
\end{figure*}

In Figure \ref{supp_fig1}-\ref{supp_fig13}, we exhibit the results of several example images with different densities from sparse, medium to dense. As seen from these results, our P2PNet achieves impressive localization and counting accuracy under various crowd \ywu{density conditions}.

Additionally, from these qualitative results, we also find that P2PNet may fail on some extreme large heads and gray images (old photos). But similar failure cases could also be found in other top
methods, such as ASNet (CVPR’20) \cite{jiang2020attention}, AMSNet (ECCV’20) \cite{hu2020count}, SDANet (AAAI’20) \cite{miao2020shallow}, \textit{etc}. Fortunately, these might be alleviated to some extent by adding more relevant training data.

\end{document}